\documentclass[conference]{IEEEtran}
\IEEEoverridecommandlockouts
\usepackage{cite}
\usepackage{amsmath,amssymb,amsfonts}
\usepackage{algorithmic}
\usepackage{graphicx}
\usepackage{textcomp}
\usepackage{xcolor}
\usepackage{algorithm}
\usepackage{algorithmic}
\usepackage{amsmath}
\usepackage{graphics}
\usepackage{epsfig}
\usepackage{amssymb}
\usepackage{url}
\usepackage{makecell}
\usepackage{amsmath}
\usepackage{subfigure}
\usepackage{balance}

\def\BibTeX{{\rm B\kern-.05em{\sc i\kern-.025em b}\kern-.08em
		T\kern-.1667em\lower.7ex\hbox{E}\kern-.125emX}}
\begin{document}
	
	\title{
		DNN-Driven Compressive Offloading for Edge-Assisted Semantic Video Segmentation 
	}
	\author{\IEEEauthorblockN{Xuedou Xiao{${^\dag}{^\ddag}$}, Juecheng Zhang{${^\dag}{^\ddag}$}, Wei Wang{${^\dag}{^\ast}$}, Jianhua He{${^\S}$}, Qian Zhang{${^\P}$}}\IEEEauthorblockA{{${^\dag}$}School of Electronic Information and Communications, Huazhong University of Science and Technology\\ {${^\S}$}School of Computer Science and Electronic Engineering, Essex University\\ {${^\P}$}Department of Computer Science and Engineering, Hong Kong University of Science and Technology\\Email: \{xuedouxiao, juechengzhang, weiwangw\}@hust.edu.cn, j.he@essex.ac.uk, qianzh@cse.ust.hk}
	\thanks{${^\ast}$The corresponding author is Wei Wang (weiwangw@hust.edu.cn).} 
	\thanks{${^\ddag}$Both authors have equal contribution.} 
	}	
	\maketitle
	
	\begin{abstract}
		Deep learning has shown impressive performance in semantic segmentation, but it is still unaffordable for resource-constrained mobile devices. While offloading computation tasks is promising, the high traffic demands overwhelm the limited bandwidth. Existing compression algorithms are not fit for semantic segmentation, as the lack of obvious and concentrated regions of interest (RoIs) forces the adoption of uniform compression strategies, leading to low compression ratios or accuracy. This paper introduces STAC, a DNN-driven compression scheme tailored for edge-assisted semantic video segmentation. STAC is the first to exploit DNN's gradients as spatial sensitivity metrics for spatial adaptive compression and achieves superior compression ratio and accuracy. Yet, it is challenging to adapt this content-customized compression to videos. Practical issues include varying spatial sensitivity and huge bandwidth consumption for compression strategy feedback and offloading. We tackle these issues through a spatiotemporal adaptive scheme, which (1) takes partial strategy generation operations offline to reduce communication load, and (2) propagates compression strategies and segmentation results across frames through dense optical flow, and adaptively offloads keyframes to accommodate video content. We implement STAC on a commodity mobile device. Experiments show that STAC can save up to 20.95\% of bandwidth without losing accuracy, compared to the state-of-the-art algorithm.

	\end{abstract}
	
	
	\section{Introduction}
	Recent years have witnessed the pervasive use of semantic segmentation in mobile vision systems, ranging from smartphones, tablets, drones, autonomous vehicles to augmented-reality headsets. Propelled by semantic segmentation, self-driving cars~\cite{cars} and drones can obtain a pixel-level understanding of surroundings. Google portrait mode~\cite{google} and YouTube stories~\cite{youtube} can enrich the background settings. Virtual try-on and make-up apps~\cite{try-on} become a reality. 
	However, the wide application of semantic segmentation is attributed to the breakthrough of deep learning technologies (e.g., deep neural networks (DNNs)). What has plagued researchers is the conflict between the growing complexity of DNN and the limited computing resources of mobile vision systems. Driven by mobile edge computing (MEC), a promising solution to alleviate this conflict is to offload DNN inference tasks to the edge servers.

	However, with the explosion of vision-based devices and applications, bandwidth limitation becomes the most pressing challenge that curbs the development of edge-assisted video analytics. Worse still, semantic segmentation does not have obvious and concentrated regions of interest (RoIs) like object detection. Therefore, existing compression algorithms/codecs~\cite{xie2019source,wallace1992jpeg,wiegand2003overview} need use a uniform and fixed compression strategy to the entire image/frame plane, such as JPEG, H.264 and GRACE. In this case, the end devices must reduce the compression ratio of the entire image/frame plane to ensure DNN accuracy, which leads to large additional bandwidth consumption and hinders the scalability of semantic segmentation. Thus, a compression scheme dedicated to semantic segmentation is urgently needed to tackle this problem. 
	
\begin{figure}[t]
		\centerline{\includegraphics[width=0.48\textwidth]{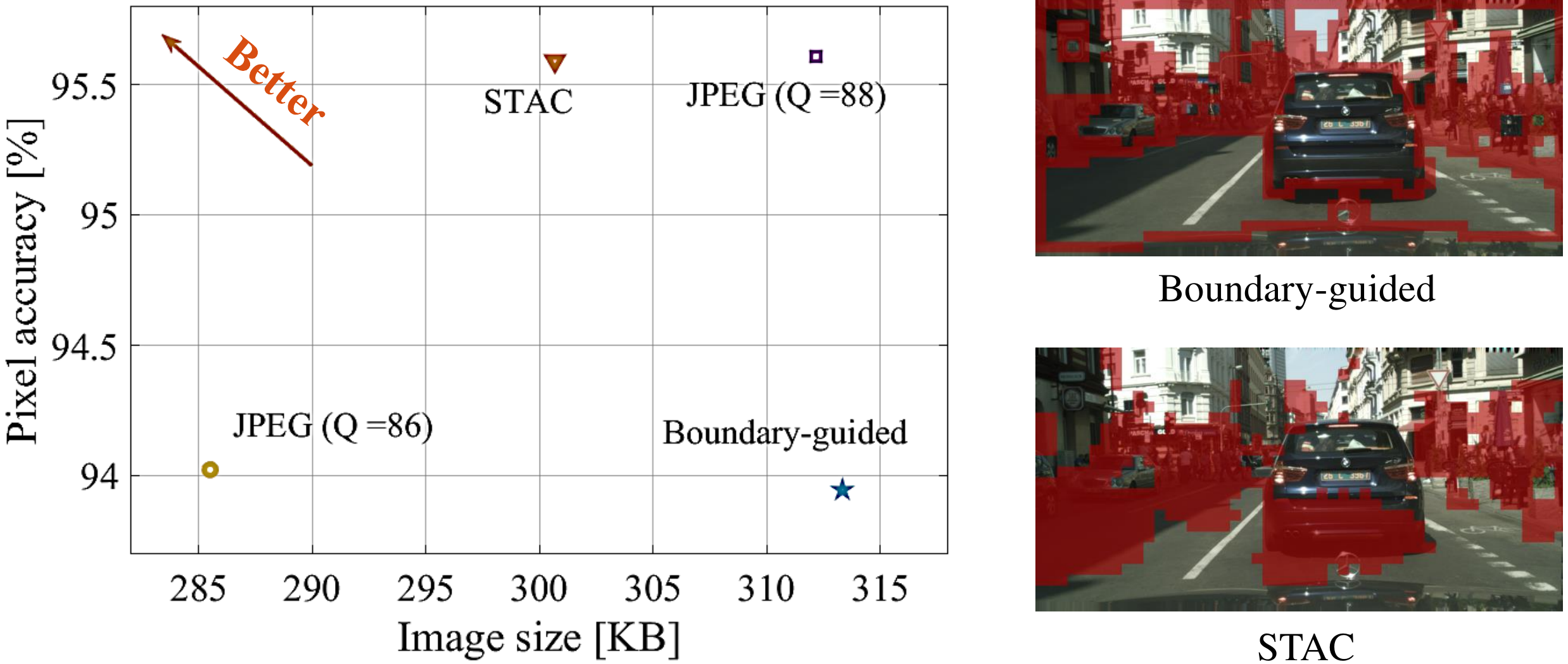}}\vspace{-0.2cm}
		\caption{A comparison of boundary-guided compression, JPEG and STAC.}\vspace{-0.4cm}
		\label{intro}
	\end{figure}

	One question to be asked is if regions around segment boundaries would be RoIs. Our experimental results shown in Fig.~\ref{intro}, however, reveal the following phenomenon: the boundary-guided compression strategy that allocates a lower compression ratio to regions around boundaries and a higher compression ratio to other regions leads to lower accuracy than the uniform compression strategy JPEG (Q=88) with comparable image size. Similarly, the boundary-guided compression strategy also results in a larger image size than JPEG (Q=86) with comparable accuracy. The red areas in the right images of Fig.~\ref{intro} indicate regions with a lower compression ratio.


	
	
\begin{figure*}[t]
	\centerline{\includegraphics[width=0.99\textwidth]{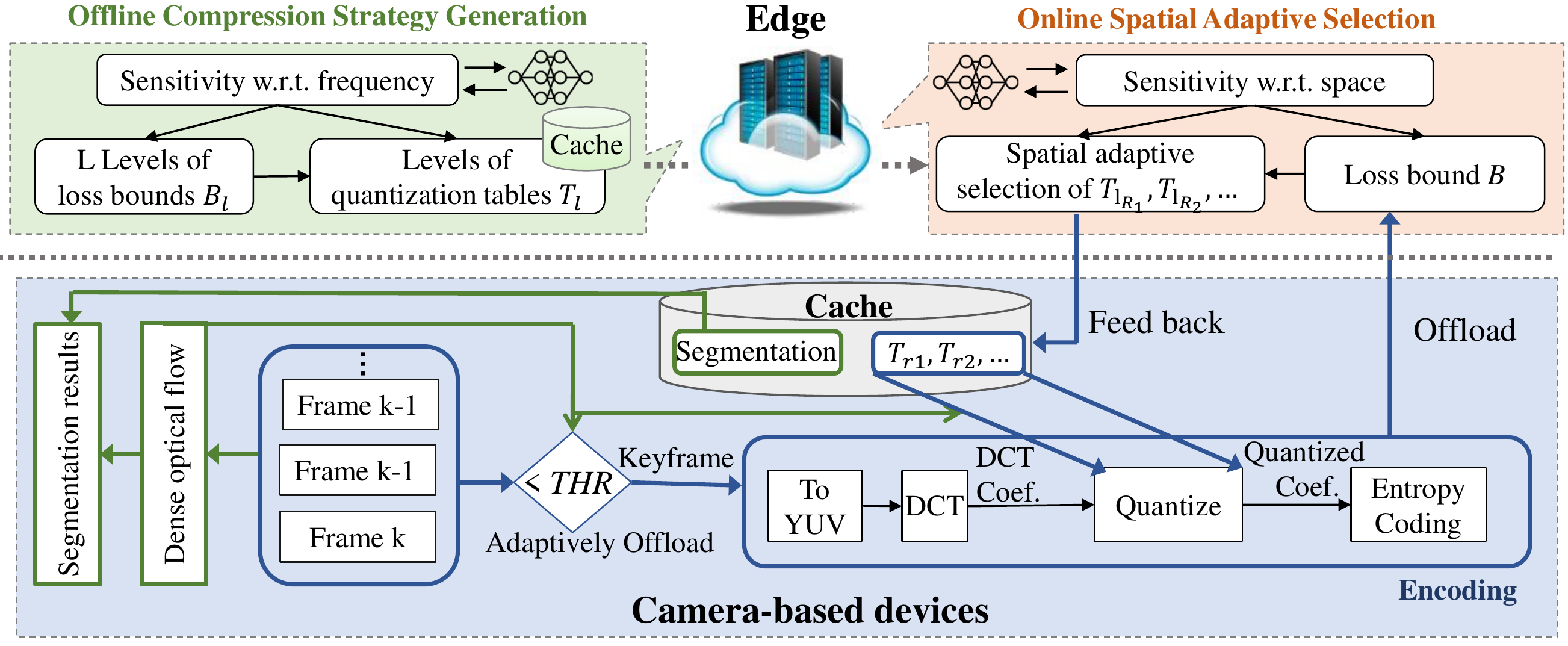}}\vspace{-0.1cm}
	\caption{System architecture of STAC.}\vspace{-0.3cm}
	\label{model}
\end{figure*}

	To accurately detect RoIs, this paper takes the first step to design a DNN-driven spatiotemporal adaptive compression scheme for semantic video segmentation. We call it STAC, which can guarantee the DNN inference accuracy while minimizing the bandwidth consumption. The core idea is that STAC utilizes DNN's gradients as nonuniform spatial sensitivity metrics of offloaded frames, to generate and feed back spatial adaptive compression strategies to end devices in real time. Specifically, the gradient quantifies the change in DNN loss function caused by each pixel undergoing compression.
	We can customize the compression strategy according to the maximum compression ratio that each pixel can withstand to achieve superior compression ratio and accuracy.


	To achieve the spatial adaptive compression, STAC entails some technical challenges.
	One key obstacle is that the spatial adaptive compression strategy is customized to the offloaded frame. Therefore, this compression strategy requires to be continuously fed back and offloaded in real time for the compression and decompression of subsequent frames. Doing so, however, consumes huge bandwidth and makes compression strategy inefficient. For example, if the offloaded frame has $M$-pixel, the feedback and offloading of the compression strategy will be up to $3\times M$ bytes, even much larger than the size of fame. To tackle this problem, we propose a spatial adaptive scheme, which no longer targets the compression strategy design for each pixel, but uses regions of frames as the basic unit for spatial adaptation. In addition, 
	STAC takes a part of the online strategy generation offline by designing different levels of compression strategies based on average gradients. Therefore, only the levels of compression strategies selected by each region need to be fed back and offloaded.

	Another key obstacle is that the customized compression strategy is not adaptive to changing video content. 
	Since the spatial sensitivity changes with video content, the compression strategy generated by the previously offloaded frame may not adapt to new frames, which have different spatial sensitivities. In other words, the compression strategy can neither be applied to already offloaded frames nor subsequent frames, making it unsuitable for video applications. To tackle this problem, We propose a temporal adaptive scheme, which extracts dense optical flow to propagate both the compression strategy
	and segmentation results across frames. This is achieved based on the temporal consistency of videos. In addition, an adaptive offloading mechanism is designed to determine whether the current frame is offloaded to update semantic segmentation and compression strategies.

	We implement STAC on a portable and small form factor Intel NUC Kit NUC7i5DNHE and prototype it on JPEG for offloaded frames. We compare STAC with a range of compression algorithms/codecs, including GRACE~\cite{xie2019source}, H.264 and JPEG. The experiments cover 2 semantic segmentation datasets with total video length up to 1115 seconds and 3 target DNN models. Evaluation results show that STAC achieves promising performance. 
	Compared to the state-of-the-art algorithm GRACE, STAC is able to save 20.95\% of bandwidth without any loss of accuracy, while improving accuracy (mIoU) up to 0.7\% when the bandwidth consumption is the same. 

	The main contributions are summarized as follows.
	\begin{itemize}
		\item STAC is the first to exploit DNN's gradients as the nonuniform spatial sensitivity metrics for spatial adaptive compression, which minimizes the bandwidth consumption without compromising the accuracy of edge-assisted semantic video segmentation.
		
		\item We propose a temporal adaptive scheme that enables the propagation of both  segmentation results and compression strategies across frames, and adaptively offloads keyframes to accommodate changing video content.
		
		\item We implement and evaluate STAC on a commodity mobile device. Experimental results confirm	the superiority of STAC compared to several baselines.
		

	\end{itemize}

	The rest of this paper is organized as follows. Section~\ref{overview} overviews the system design of STAC. Section~\ref{section3} elaborates on the spatial adaptive compression strategy and Section~\ref{temporal} introduces the temporal adaptive scheme. Finally, we present our implementation details in Section~\ref{implementation} and our experimental results in Section~\ref{evaluation}. Section~\ref{related} reviews the related works, followed by the conclusion in Section~\ref{conclusion}.


	

	


	\section{System Overview} \label{overview}

	Fig.~\ref{model} illustrates the system architecture of STAC, which is split into three stages including offline compression strategy generation (Section~\ref{offline}), online spatial adaptive selection~(Section~\ref{online}) and temporal adaptation~(Section~\ref{temporal}).
	
	At the stage of offline compression strategy generation, STAC first measures DNN's average gradients (i.e., sensitivity) w.r.t. different frequency of discrete cosine transform (DCT) coefficients at the edge, inspired by \cite{xie2019source}. Then, depending on the accuracy requirement, STAC configures $L$ levels of upperbounds $\{B_l\}_{l=1}^{L}$ of the allowed DNN loss increments. Next, combining the gradients and upperbound $B_l$, STAC can generate $L$ levels of block-wise quantization tables $\{T_l\}_{l=1}^{L}$ accordingly, each of which corresponds to a different compression ratio and $B_l$. 
	
	At the stage of online spatial adaptive selection, STAC generates spatial adaptive compression strategy online at the edge. Specifically, STAC measures DNN's gradients w.r.t. each DCT coefficient of the whole offloaded keyframe, and calculates the worst-case loss increment for each region $R_r \ (r\in\{1,2,...,r_{max}\})$ using different levels of $T_l$. Depending on the required upperbound $B$, the best $T_{l_{R_r}}$ for each region $R_r$ can be selected, which has a worst-case loss increment closest to $\frac{B}{r_{max}}$. Therefore, only levels $\{l_{R_r}\}_{r = 1}^{r_{max}}$ of selected $\{T_{l_{R_r}}\}_{r = 1}^{r_{max}}$ for all regions $\{R_r\}_{r = 1}^{r_{max}}$ and the segmentation results are fed back to the end devices. 
	
	The temporal adaptation is entirely performed online at the end devices.	
	STAC caches the compression strategy and semantic segmentation results fed back from the edge for future use. Then, using the dense optical flow computed by Dense Inverse Search (DIS)~\cite{kroeger2016fast}, STAC propagates prior semantic segmentation to later frames, and propagates the spatial adaptive compression strategy to later keyframes for encoding and offloading. The mechanism of deciding whether a frame is offloaded or not is introduced in Section~\ref{offloading}.


	\section{Leveraging Nonuniform Spatial Sensitivity} \label{section3}
	In this section, we first give a brief introduction to existing video/image codecs and  semantic video segmentation. Then, we elaborate on the generation of spatial adaptive compression strategy in STAC, including the offline stage and online stage.
	
	\subsection{Primers}
	
	\subsubsection{Video/Image codec} The most common steps of image codecs (e.g., JPEG) and intra-frame compression in video codecs (e.g., H.264) are shown in the bottom right part of Fig.~\ref{model}. The encoder first converts an RGB image to YUV format, and utilizes discrete cosine transform (DCT) to obtain frequency-based DCT coefficients. Existing codecs like JPEG, H.264, etc., generally use block-wise DCT (e.g., 8 × 8), instead of applying DCT to the whole image. All Y\&U\&V components are operated on separately. The next stage is to quantize DCT coefficients for each block by a quantization table $T=\{q_1, q_2,..., q_N\}$, where each $q_n$ represents the quantization step on DCT coefficients of $n_{th}$ frequency, $N$ (e.g., 8 × 8) is the number of both pixels and DCT coefficients in a block. 
	After ZigZag scanning of the quantized coefficients from low frequency to high frequency, the entropy encoding like Huffman is used to further compress the quantized coefficients such as the consecutive '0' and consecutive '1'.

	\begin{figure}[t]
		\centerline{\includegraphics[width=0.5\textwidth]{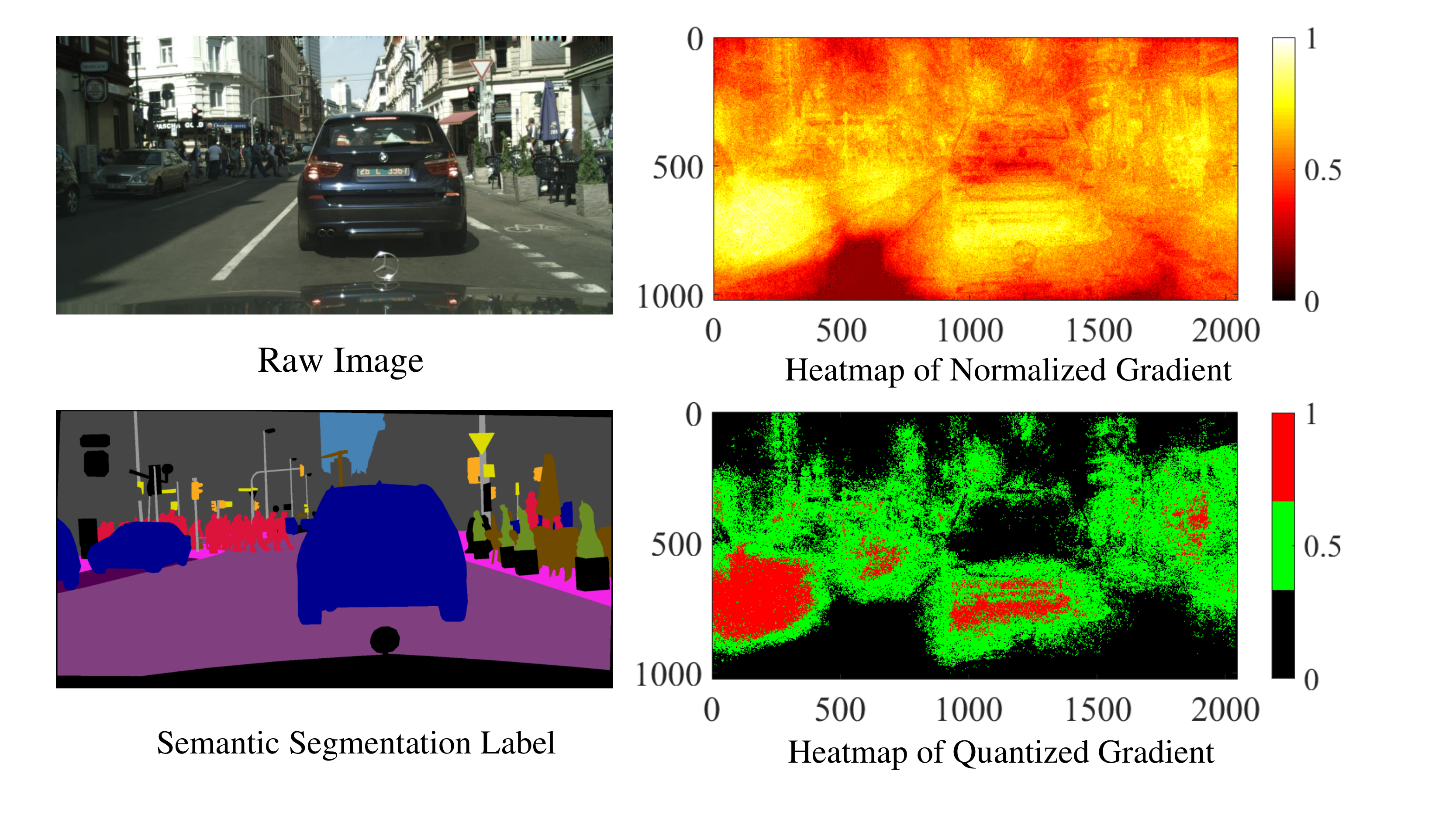}}
		\vspace{-0.1cm}\caption{The heatmaps of DNN's gradients on a random raw image demonstrate the nonuniform spatial sensitivity.}\vspace{-0.2cm}
		\label{zong_heatmap}\vspace{-0.2cm}
	\end{figure}
	
	\subsubsection{DNN-based semantic video segmentation}
	Two approaches are often used in semantic video segmentation. One is to input consecutive frames into DNN models~\cite{mahasseni2017budget,shelhamer2016clockwork,li2018low}, and the other is to perform DNN-based semantic image segmentation on each frame~\cite{xie2019source} or keyframes~\cite{xu2018dynamic,paul2020efficient,zhu2019improving}. The DNN models that take consecutive frames as input~\cite{mahasseni2017budget,shelhamer2016clockwork,li2018low} generally result in a large number of network parameters and are computationally intensive. They are impractical for edge-assisted video analytics like ours, as they require to offload consecutive frames, leading to severe bandwidth consumption. Therefore, we adopt the second approach and only keyframes are offloaded to minimize the bandwidth consumption. To further obtain continuous semantic segmentation for videos, we leverage the temporal consistency and extract the dense optical flow to propagate both compression strategy and semantic segmentation across frames, which is detailed in Section~\ref{temporal}.

	\subsection{Offline Compression Strategy Generation}\label{offline}
	We elaborate on the offline compression strategy generation from the perspective of the following two questions.
	\subsubsection{Why we should generate compression strategy offline}\label{offline-1}
	Consider a DNN with loss function $Q$ and an $M$-pixel input image/frame represented as $\textbf{x} = \{x_1, x_2,...,x_M\}$. We follow the approach in GRACE~\cite{xie2019source} to obtain the DNN's gradient $g_{x_i}$ w.r.t. each pixel $x_i$ by calculating the partial derivative of $Q$ with respect to $x_i$, i.e., $g_{x_i} = \frac{\delta Q}{\delta x_i}$. 
	We plot in Fig.~\ref{zong_heatmap} the heatmap of $g_{x_i}$ on a random raw image, which verifies the nonuniform spatial sensitivity. For more clarity, we quantize the gradient and it can be noticed that the most sensitive regions are often not the boundaries. Thus, a specialized spatial adaptive compression strategy is needed for semantic segmentation.
	According to the total differential equation, when all $\Delta x_i$ are very small, the loss change $\Delta Q$ of DNN can be modeled as $\Delta Q = \sum_{i = 1}^M g_{x_i}  \Delta x_i$. 
	Since quantization is one of the major lossy compression techniques during encoding, we convert DNN's gradients w.r.t. pixels $g_{\textbf{x}}$  to DNN's gradients w.r.t. DCT coefficients $g_{\textbf{s}}$. Therein, \textbf{s} is a vector of DCT coefficients $\{s_1, s_2, ..., s_M\}$ and the corresponding gradient is $g_{s_i}$. 
	Thus, $\Delta Q$ can also be expressed as 
	\begin{equation}
		 \Delta Q =\sum_{i = 1}^M g_{x_i}  \Delta x_i =  \sum_{i = 1}^M g_{s_i}  \Delta s_i, 
	\end{equation}
	where $\Delta s_i = s_i - \lfloor \frac{s_i }{q_{s_i}} \rceil q_{s_i}$ denotes the quantization error and $\left|  \Delta s_i \right| \leq \frac{q_{s_i}}{2}$. 
	Therefore, we set $\Delta Q_{max} = \sum_{i = 1}^M |g_{s_i}|\frac{q_{s_i}}{2}$, which represents the worst-case loss increment caused by quantization steps of the whole image $\{q_{s_1}, q_{s_2},...,q_{s_M}\}$. When we set a upperbound $B$ of the allowed loss increment to guarantee the DNN inference accuracy, the following constraint is satisfied: $\sum_{i = 1}^M |g_{s_i}|\frac{q_{s_i}}{2} \leq B$. The objective becomes to minimize the bandwidth consumption under this constraint by designing proper $\{q_{s_1}, q_{s_2},...,q_{s_M}\}$, denoted as $\arg\min\limits_{\textbf{q}} \sum_{i = 1}^M \log_2{\left|  \frac{s_i}{q_{s_i}} \right|} $. When $d_{s_i} = |g_{s_i}|\frac{q_{s_i}}{2}$, this problem is simplified to
	\begin{align}
		\arg\min\limits_{\textbf{q}} \sum_{i = 1}^M \log_2{\left| \frac{s_i}{q_{s_i}} \right|} =  \arg\max\limits_{\textbf{q}} \prod_{i = 1}^M d_{s_i} \ \
		s.t. \sum_{i = 1}^M d_{s_i}\leq B,
	\end{align}
	The optimal solution is when all $\{d_{s_i}\}_{i = 1}^M$ are equal and $d_{s_i} =\frac{B}{M}$. Therefore, the ideal design of quantization step $\{q_{s_1}, q_{s_2},...,q_{s_M}\}$ for all DCT coefficients in an image/frame is $q_{s_i} = \frac{2B}{M \left| g_{s_i} \right|} $, which minimizes the file size while not exceeding the upperbound $B$ to guarantee the inference accuracy.
	
	
	\begin{algorithm} [t]
		\renewcommand{\algorithmicrequire}{\textbf{Input:}}
		\renewcommand{\algorithmicensure}{\textbf{Output:}}
		\caption{Online Spatial adaptive compression strategy. } 
		\label{spatial} 
		\begin{algorithmic} [1]
			\REQUIRE $\{q_1^l,q_2^l,...,q_N^l\}_{l = 1}^L$ – $L$ quantization table levels\\
			$B$ – Upperbound of the allowed DNN loss increment \\
			$Q$ – DNN loss function \\
			$\{s_i\}_{i = 1}^{M}$ – DCT coefficients of the offloaded frame \\
			$\{R_r\}_{r = 1}^{r_{max}}$ – Regions of the offloaded frame
			\ENSURE $\{l_{R_r}\}_{r = 1}^{r_{max}}$ – Levels of quantization tables selected by regions  
			\FOR{ $i \gets 1 $ to $M$} 
			\STATE $g_{s_i} = \frac{\delta Q}{\delta s_i}$  
			\ENDFOR
			\FOR{ $r \gets 1 $ to $r_{max}$}
			\STATE  $min = +\infty $ 
			\FOR{ $l \gets 1 $ to $L$}
			\STATE
			$\Delta Q^l_{max, R_r} \gets \sum_{s_i \in R_r} \left| g_{s_i} \right|\times \frac{q^l_{s_i\rightarrow n}}{2}$  \\
			\IF{ $\left| Q^l_{max, R_r}-\frac{B}{r_{max}}\right| <min $}
			\STATE $l_{R_r} = l$
			\STATE  $min = \left| Q^l_{max, R_r}-\frac{B}{r_{max}}\right|$
			\ENDIF
			
			\ENDFOR
			\ENDFOR		
			
		\end{algorithmic} 
	\end{algorithm}
	
	
	This compression strategy is generated online at the edge, and thus requires constant feedback to the end device for adoption. In addition, when the compression strategy is propagated and applied to subsequent frames (detailed in Section~\ref{temporal}), it also needs to be offloaded for decoding.
	This heavy communication load makes the design impractical. For example, supposing a frame has $M = 2048\times1024$ pixels, the total number of DCT coefficients reaches $3\times2048\times1024$. 
	Therefore, up to 6M quantization steps, i.e., 6MB data need to be transmitted. This is even larger than the raw frame. When frames are offloaded frequently, this communication load of compression strategy will severely consume bandwidth and make the compression strategy ineffective.
	
	\subsubsection{What this offline compression strategy is}
	To ease the communication load of compression strategy, the most effective way is to take a  part of the compression strategy generation operations offline to reduce the online feedback and offloading content. Many compression strategies such as JPEG, H.264 and GRACE~\cite{xie2019source} are generated offline. However, all of them 
	apply fixed block-wise quantization tables $T = \{q_1,q_2,...,q_N\}$ to each block (e.g., $N = 8\times 8$), which ignore the nonuniform spatial sensitivity. Although the quantization table $T$ in GRACE is also generated by DNN's gradients w.r.t. DCT coefficients, these gradients $\{g_n\}_{n = 1}^{N}$ are average values across all blocks, representing DNN's sensitivity to each frequency component $n$. The spatial sensitivity is still not leveraged.
	However, as shown in Fig.~\ref{zong_heatmap}, DNN's gradients are varying across  regions and blocks. 
	
	Thus, our goal becomes how to design offline compression strategies that can support STAC to exploit the nonuniform spatial sensitivity online and reduce the communication load of compression strategy. The idea is to generate $L$ levels of block-wise quantization tables $\{T_l\}^L_{l=1}$ offline, which are then available online and selected by different regions of the frame according to their sensitivities. Each $T_l$ represents a different compression ratio and is determined by different upperbounds $\{B_l\}_{l=1}^{L}$. In this way, STAC no longer transmits the overall compression strategy, but simply the levels $\{l\}_{l=1}^L$ of offline quantization tables selected by different regions.
	
	The detailed steps of generating multiple levels of $\{T_l\}^L_{l=1}$ are as follows: (1) First, we also calculate DNN's average gradients $\{g_n\}_{n=1}^{N}$ w.r.t. different frequency components of DCT coefficients, to fully incorporate the advantage of GRACE. (2) Depending on the accuracy requirement, we search for the maximum upperbound $B$ that satisfies the requirement. The searching  process includes traversing upperbound $B$ values, calculating their corresponding block-wise $T= \{q_{1},q_{2},...,q_{N}\}, q_n = \frac{2B}{M\left|g_n\right|}$, and testing the inference accuracy when applying these quantization tables to images/frames.
	(3) We then configure $L$ levels of upperbound $\{B_l\}_{l = 1}^L$ around $B$. Accordingly, $L$ levels of $\{T_l\}_{l = 1}^L= \{q_{1}^l,q_{2}^l,...,q_{N}^l\}_{l = 1}^L, q_n^l = \frac{2B_l}{M\left|g_n\right|}$ are generated offline. We set $L$ to 16 and the reason is explained in Section~\ref{configurations}.
	It is noteworthy that these offline quantization tables are only customized to DNN, which can be obtained based on frames that are not within the test set. 
	
	\subsection{Online Spatial Adaptive Selection}\label{online}

	Based on the generated $L$ levels of quantization tables offline, STAC is able to select  proper $T_l$ for different regions of the offloaded frames adaptively. 
	For example, for a region with low sensitivity (i.e., low gradient), $T_l$ with high compression ratio is given priority. Conversely, for a region with high sensitivity, $T_l$ with low compression ratio is preferred. For simplicity, we divide the frame into $r_{max}$ regions, each of which contains the same number of blocks. We use the region as the basic unit for $T_l$ selection, which further reduces the communication load of compression strategies. 
	Based on the evaluation results in Section~\ref{configurations}, $r_{max}$ is set to $\lceil \frac{M}{N\times3\times3} \rceil$.
	
	The steps of online quantization table selections for regions $\{R_r\}_{r = 1}^{r_{max}}$ are as follows: (1) 
	First, STAC measures DNN's gradients $\{g_{s_i}\}_{i = 1}^M$ w.r.t. all DCT coefficients $\{s_i\}_{i =1}^M$ of the offloaded frame. (2) Then, STAC calculates the worst-case loss increment for each region $R_r$ under each level of quantization table $T_l = \{q^l_1,q^l_2,...,q^l_N \}$, which is denoted as $\Delta Q_{max,R_r}^l = \sum_{s_i \in R_r} \left| g_{s_i} \right|\times \frac{q^l_{s_i\rightarrow n}}{2}$. Therein,  $s_i\rightarrow n$ denotes the frequency component $n$ of DCT coefficient $s_i$.
	(3) Next, for each region $R_r$, STAC selects the quantization table $T_{l_{R_{r}}}$, which has the worst-case loss increment $\Delta Q_{max,R_{r}}^l$ closest to the upperbound $\frac{B}{r_{max}}$ assigned to this region. $\frac{B}{r_{max}}$ represents that the total upperbound $B$ is uniformly assigned to each region based on the number of pixels it contains. $B$ is decided offline according to the accuracy requirement. (4) The level $l_{R_{r}}$ of quantization table $T_{l_{R_{r}}}$ selected by each region $R_r$ is fed back online. Following these steps, we summarize the online spatial adaptive 
	compression algorithm in Alg.~\ref{spatial}. 
	
	Recall that we have theoretically derived in Section~\ref{offline-1} that the compression ratio is maximum when each $q_{s_i} = \frac{2B}{M\left| g_{s_i} \right|}$. Due to the heavy communication load, STAC no longer targets the quantization step for each DCT coefficient, but only controls the quantization table $T_{l_{R_r}}$ for each region $R_r$, so that the $\Delta Q_{max,R_r}^l$ is closest to the upperbound $\frac{B}{r_{max}}$ assigned to this region. 
	However, we are still unclear whether this can achieve a performance gain, i.e., a higher compression ratio compared to conventional algorithms without spatial adaptation, when the accuracy is the same (i.e., same upperbound $B$). Therefore, we prove it theoretically in this section, followed by an experimental verification in Section~\ref{compare}. The problem can be formulated as follows.
		
	\begin{figure}[t]
		\centerline{\includegraphics[width=0.46\textwidth]{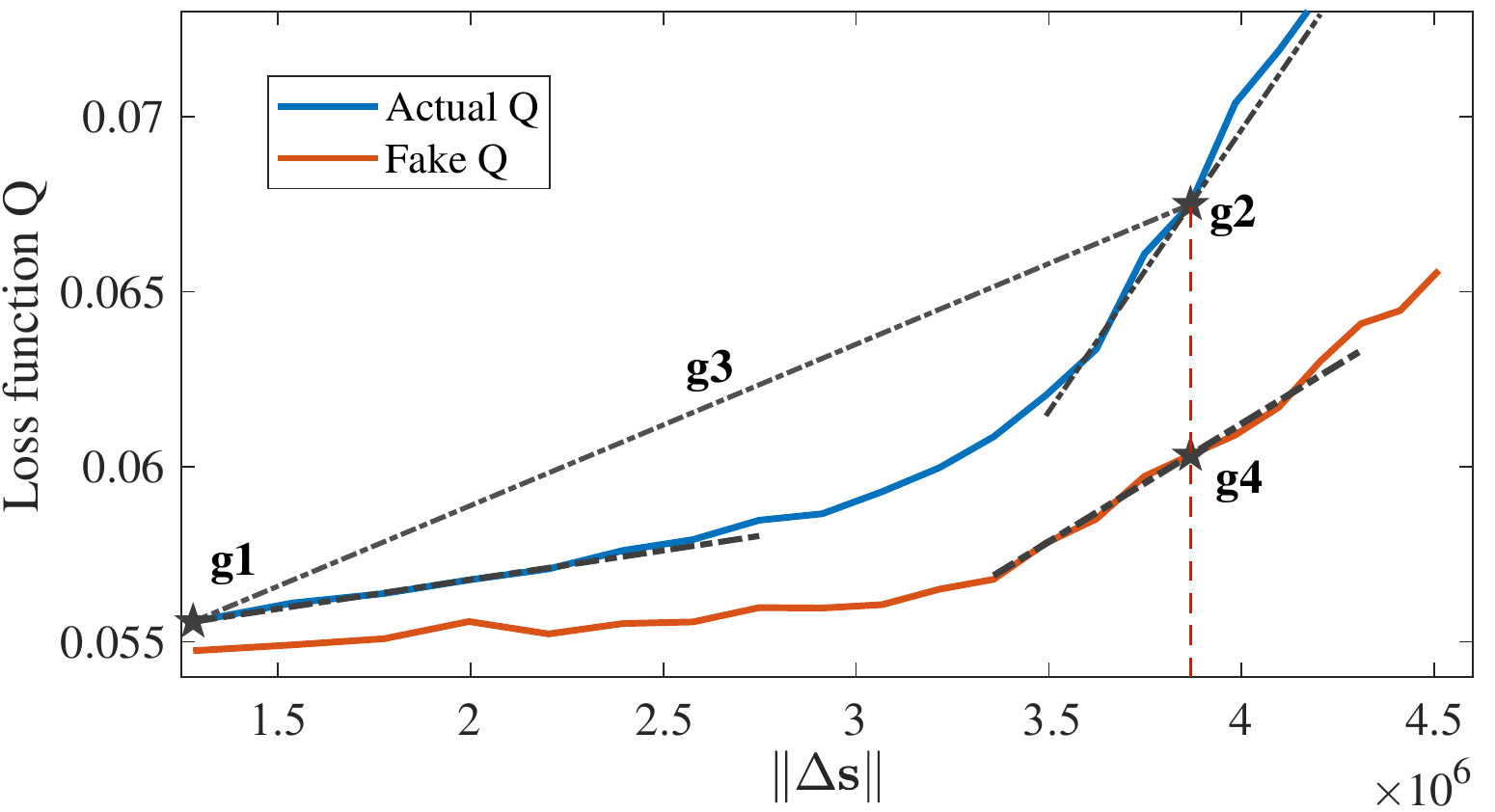}}
		\vspace{-0.2cm}\caption{Fake $Q$ vs. actual $Q$ when compression ratio increases.}\vspace{-0.2cm}
		\label{xielv}\vspace{-0.2cm}
	\end{figure}
	
	
	
	\textit{Problem: When $ \sum_{i = 1}^M d_{s_i}= B$ (C1) and each $d_{s_i} \geq 0$  (C2), can $\mathbb{E}(\prod_{i = 1}^M d_{s_i}) $ under the constraint $\sum_{s_i \in R_1}d_{s_i} = \cdots= \sum_{s_i \in  R_{r_{max}} } d_{s_i} = \frac{B}{r_{max}} $ (C3) be larger than $\mathbb{E}(\prod_{i = 1}^M d_{s_i})$ without this constraint?}
	
	Proof: (1) With the constraints C1 and C2 but without C3, 
	the joint probability distribution is 
	\begin{small} 
	\begin{align}
		f(d_{s_1},...,d_{s_M}) = \frac{1}{\int...\int_{C1,C2} d (d_{s_1})... d (d_{s_M})} = \frac{(M-1)!}{B^{M-1}}.
	\end{align}
	\end{small}
	Then, the expectation value can be estimated by
	\begin{small} 
	\begin{align}
		\mathbb{E}_1(\prod_{i = 1}^M& d_{s_i}) =\frac{(M-1)!}{B^{M-1}} \int\cdots\int_{C1,C2} d_{s_1}...d_{s_M} \ d (d_{s_1})... d (d_{s_M}) \notag\\ 
		&= \frac{(M-1)!}{B^{M-1}}  \frac{B^{2M-1}}{(2M-1)!} = \frac{B^M}{(2M-1)(2M-2)...M}.
	\end{align}
	\end{small}
	(2) When all these constraints C1, C2 and C3 are satisfied, the expectation value can be calculated by
	\begin{small} 
		\begin{align}
			\mathbb{E}_2(\prod_{i = 1}^M d_{s_i}) = [\mathbb{E}_1(\prod_{s_i \in R_{1}}d_{s_i})]^{r_{max}}.
		\end{align}
	\end{small}
	For simplicity, we assume each region $R_r$ has approximate $\frac{M}{r_{max}}$ pixels. 
	Thus, the expectation value can be estimated as
	\begin{small} 
		\begin{align}
			\mathbb{E}_2(\prod_{i = 1}^M d_{s_i}) =\left[\frac{(\frac{B}{r_{max}})^{\frac{M}{r_{max}}}}{[(2\frac{M}{r_{max}}-1)(2\frac{M}{r_{max}}-2)...\frac{M}{r_{max}}]}\right]^{r_{max}}.
		\end{align}
	\end{small}
	Thus, we can obtain 
	\begin{small} 
		\begin{align}
			\frac{\mathbb{E}_1}{\mathbb{E}_2} = \frac{[(2M-r_{max})(2M-2r_{max})...M]^{r_{max}} }{(2M-1)(2M-2)...M} \textless  1.
		\end{align}
	\end{small}
	From these equations, we theoretically verify that adding constraint C3 does lead to performance improvement under the same upperbound $B$ of loss increment, which provides theoretical justifications for our approach.

	One concern is whether it is necessary to first offload a raw frame to calculate the ground truth for gradients.
	We argue that the fake gradient computed by the compressed frame is even more appropriate than the actual gradient computed by the raw frame.
	The fake gradient is computed as follows: (1) We treat the DNN inference result of the compressed frame as the fake ground truth/label; (2) We calculate the fake loss function $Q$ between the probability vector of DNN output and the fake label; (3) The fake gradient is calculated. We have tested both actual $Q$ and fake $Q$. The results are shown in Fig.~\ref{xielv}, with x-axis $\Vert \Delta \textbf{s}  \Vert$ represents the total quantization errors as compression ratio increases, i.e., $\Vert \Delta \textbf{s}  \Vert = \left|\Delta s_1 \right|+...+\left|\Delta s_M \right|$. Both actual and fake $Q$ are concave, but fake $Q$ increases slowly. The reason is that the fake label is closer to the DNN output vector than the actual label. For example, assuming that DNN output is $(0.6, 0.4, 0)$ and the actual label is $(0,1,0)$, the fake label however can be computed as $(1,0,0)$, which results in lower loss value. $g1$ and $g2$ in Fig.~\ref{xielv} respectively denote the actual gradients of the raw frame (nearly) and the compressed frame. $g3$ represents the actual speed of loss increment under compression. Both $g1$ and $g2$ are very different from $g3$, while $g4$ that is computed by fake $Q$ and compressed frames is more similar to $g3$, which is more appropriate to calculate the quantization strategy.


	\section{Leveraging Temporal Consistency}\label{temporal}
	In this section, we propose a temporal adaptive scheme that propagates both segmentation results and spatial compression strategies across frames, and adaptively offloads keyframes according to video content changes.
	We first briefly introduce the dense optical flow used for propagation, and then dive into the cross-frame propagation and the adaptive offloading.
	\subsection{A Primer on Dense Optical Flow}
	Optical flow is defined as the motions of pixels between adjacent frames. 
	Different from the sparse optical flow that only focuses on interesting features, dense optical flow provides flow vectors of the entire frame, up to one flow vector per pixel. Recent works~\cite{paul2020efficient,zhu2019improving,xu2018dynamic} have extracted dense optical flow to propagate semantic segmentation across frames. STAC also adopts DIS~\cite{paul2020efficient}, which is state-of-the-art in terms of computational efficiency on CPU, to propagate both compression strategy and semantic segmentation. Compared to deep optical flow methods, DIS achieves competitive accuracy and higher frame rates of 10-600~Hz on 1024$\times$436 resolution images on a single CPU core. This is fully compatible with mobile vision systems such as mid-end smartphones, tablets, AR headsets, etc. Besides, DIS operates in a coarse-to-fine fashion, giving STAC more flexibility to adjust computational costs according to the available runtime resources of mobile vision systems.


	

	\subsection{Propagation and Adaptive Offloading}\label{offloading}
	Recall that both segmentation results and spatial compression strategies are generated online at the edge based on the offloaded keyframe, and fed back to the end device. To avoid the delay caused by the whole pipeline of offloading, DNN inference and feedback, STAC caches segmentation results and compression strategies of the offloaded keyframe on the end device for future use.	
	After computing the pixel-wise optical flow, STAC propagates segmentation results from the keyframe to the current frame. If the current frame is determined to be offloaded, the compression strategies will also be propagated.
	It is noteworthy that the optical flow is only computed between adjacent raw frames to ensure accuracy. 
	
	\begin{algorithm} [t]
		\renewcommand{\algorithmicrequire}{\textbf{Input:}}
		\renewcommand{\algorithmicensure}{\textbf{Output:}}
		\caption{Temporal adaptive scheme.} 
		\label{Temporal adaptive scheme} 
		\begin{algorithmic} [1]
			\REQUIRE $\{F_{t}\}_{t = 1}^{T'}$ – raw frames captured on the end device \\
			$THR$ – Pre-defined threshold
			\ENSURE $\{S_{t}\}_{t = 1}^{T'}$ –  semantic segmentation results\\
			$\{F_{{kt}_{j}}\}_{j = 1}^{J}$, ${kt}_{j}\in \{t\}_{t = 1}^{T'}$ –   offloaded keyframes\\
			$\{C_{{kt}_{j}}\}_{j = 1}^{J}$ – compression strategies applied to keyframes	
			\STATE ${kt}_{1} = 1$, $j = 1$
			\STATE Compress $F_1$ with a middle level quantization table
			\STATE Offload $F_1$, then generate, feed back and cache $C_1$, $S_1$
			\FOR {$t \gets 2$ to $T'$}
			\STATE Compute optical flow $\gamma_{t-1}^{t}$ between $F_{t-1}$ and $F_t$
			\STATE Propagate $S_{kt_{j}}$, $F_{kt_{j}}$ to $S_{t}$, $F_{t}'$ by $\gamma_{kt_{j}}^{kt_{j}+1},...,\gamma_{t-1}^{t}$
			
			\IF {$PSNR (F_{t}',F_{t}) \textless THR $}
			\STATE $j \gets j+1 $, $kt_j \gets t $
			\STATE Propagate $C_{{kt}_{j-1}} $ to $C_{{kt}_{j}} $ by $\gamma_{{kt}_{j-1}}^{{kt}_{j-1}+1},...,\gamma_{{kt}_{j}-1}^{{kt}_{j}}$
			\STATE Apply $C_{{kt}_{j}} $ to $F_{kt_j}$ and offload $F_{kt_j}$  
			\STATE Generate $S_{{kt}_{j}}$, $C_{{kt}_{j}}$ and feed back to update 
			\ENDIF
			\ENDFOR

		\end{algorithmic} 
	\end{algorithm}
	
	To determine whether the current frame is offloaded as a keyframe to update semantic segmentation and compression strategies, 
	we need metrics that indicate the accuracy degradation of propagated semantic segmentation. However, there is no semantic segmentation ground truth at the end device to test the accuracy. To tackle this problem, we additionally use the dense optical flow to propagate pixels of the keyframe to the current frame. The key point is that  the propagated frame and the propagated semantic segmentation have a one-to-one correspondence at each pixel, as the optical flow is the same. Thus, we compute the pixel-level similarity (such as peak-signal-to-noise ratio (PSNR)) between the propagated frame and the actual frame to evaluate the degree of distortion. Once PSNR falls below a pre-defined threshold (26~dB according to Section~\ref{configurations}), the semantic segmentation accuracy is not guaranteed and the adaptive offloading mechanism is triggered to obtain new semantic segmentation. This mechanism can autonomously adjust offloading rate according to real-time video content, ensuring the semantic segmentation accuracy.

	Since the first frame of a video does not have a corresponding spatial adaptive compression strategy, STAC applies a middle-level quantization table to uniformly compress and offload it. Then the initial spatial adaptive compression strategy is generated and fed back to end devices. Through dense optical flow, STAC propagates it to subsequent offloaded keyframes. For example, the quantization table level $l_{R_1}$ of region $R_1$ in the previous keyframe may be propagated to region $R_3$ in the current keyframe. If there exits new regions $R_{r,new}$ where the optical flow is not available, these regions 
	adopt the quantization table of the nearest region. It is similar for the propagation of semantic segmentation. The pixels without optical flow temporarily use the semantics of the nearest pixels. After offloading and feedback, the cached semantic segmentation and compression strategies will be updated. We summarize the temporal adaptive scheme in Alg.~\ref{Temporal adaptive scheme}.
	



	\section{Implementation} \label{implementation}
	
	Since only keyframes are offloaded to the edge server, we prototype STAC on JPEG codec and only change quantization tables used for pixel blocks. This does not add any extra computations to the original codec at the end device. In addition, the quantization tables are generated offline at the edge server, and the online selection of quantization tables is also running at the edge. Besides, we cache the semantic segmentation results at the end device. Once the time for all pipelines including the compression, offloading, DNN inference and feedback exceeds the frame interval, STAC will propagate the cached segmentation results to the current frame to meet the real-time requirement.
	Therefore, the only computational stress at the end device that may increase the latency is the computation of the optical flow and the propagation.

	We evaluate STAC on a portable and small form factor Intel NUC Kit NUC7i5DNHE, which includes an Intel Core i5-7300U CPU (with 2 cores, 3~MB cache, 2.6~GHz). The specification of this hardware is comparable to what is available in today's mid-end smartphones such as OnePlus 9 and Samsung Galaxy A52 5G. We deploy an edge server equipped with an Intel Core i7-5820K CPU and an Nvidia Tesla T4 GPU.
	In addition, we use Pytorch and call C++ libraries through   interfaces~\cite{ehrlich2020quantization,kroeger2016fast} to implement all steps including offline quantization table generation, online spatial adaptive selection, source compression, optical flow computation and propagation. Furthermore, we adjust parameters of DIS to enable optical flow computation and propagation to reach 17~fps (frame rate of Cityscapes dataset~\cite{cordts2016cityscapes}) in our experiment.

\begin{table}[t]
	\renewcommand\arraystretch{1.4}
	\centering
	\caption{Side-by-side Performance Comparison
		of Different Algorithms/Codecs.} \label{table1}
	\vspace{-1.5 mm}
	\setlength{\tabcolsep}{1.2mm}{
		\begin{tabular}{|c||c|c|c|c|}
			\hline  
			\makecell[c]{Comp.\\ Format}&\makecell[c]{Offloading Bit \\ Rate (Kbps)}&\makecell[c]{Accuracy \\ (mIoU)}&\makecell[c]{Offloaded Frames \\ Per Second} \\
			
			
			\hline\hline %
			STAC &  19450 & 66.16\% &  6.8 \\  \hline
			GRACE & 23120 & 66.04\% & 6.8 \\ \hline
			H.264 (CRF = 13) & 25670 & 66.22\% & 17 \\ \hline
			JPEG (Q = 96) & 29260 &  65.94\% & 6.8\\
			
			\hline
	\end{tabular}}\vspace{-1.2mm}
\end{table}

	\section{Evaluation} \label{evaluation}

	\subsection{Experimental Setup}
	\textbf{Datasets, DNN models, and Metrics.} We evaluate STAC on semantic video segmentation task using Cityscapes dataset~\cite{cordts2016cityscapes} with 2048$\times$1040 resolution and CamVid dataset~\cite{brostow2009semantic} with 720$\times$960 resolution, respectively. Since videos in Cityscapes dataset are sparsely annotated, i.e., not every frame is annotated, we only evaluate semantic segmentation accuracy on annotated video frames and repeat the test with different starting points. 

	We employ three DNN models including DRN-D-22~\cite{DRN22}, DRN-D-38~\cite{yu2017dilated} and  BiSeNet~\cite{yu2018bisenet} for semantic segmentation to evaluate whether STAC can adapt to different DNNs and achieve performance improvement. 
	We evaluate two key metrics, including the offloading bit rate (Kbps) (i.e., bandwidth consumption) and semantic segmentation accuracy. The accuracy is measured using the standard mean intersection over union (mIoU) metric
	, where IoU represents the ratio of the intersection and merge of the predicted and true classes of pixels. mIoU is the average IoU value of all classes. Thus, 
	$mIoU=100$ indicates that the semantic predictions of all pixels exactly match the true labels, while $mIoU= 0$ indicates that the semantic predictions of all pixels are wrong.


	


	\textbf{Baseline.}
	The baseline compression algorithms or codecs are summarized as follows.
		\begin{itemize}
		\item \textit{GRACE}~\cite{xie2019source} is the state-of-the-art compression algorithm for edge-assisted semantic segmentation. It measures the DNN's gradients w.r.t. frequency components of DCT coefficients to generate a quantization table that is uniformly applied to all blocks in images/frames. As GRACE offloads each frame when processing video, we combine it with the propagation of semantic segmentation (same as STAC) for comparison.
		
		\item \textit{H.264} is a well-known industry standard for video compression or codec. It includes both intra-frame compression and inter-frame compression. In order to avoid the extra latency caused by bidirectional prediction/encoding, we control the number of B-frames to 0 and only allow I-frames and P-frames. 
		
		\item \textit{JPEG} is a commonly used image codec. Similar to GRACE, we also combine it with the propagation of semantic segmentation (same as STAC) for comparison.
	\end{itemize}

	\begin{figure}
	\centering
	\subfigure[vs. GRACE on DRN-D-22]{
		\includegraphics[width=0.48\linewidth]{"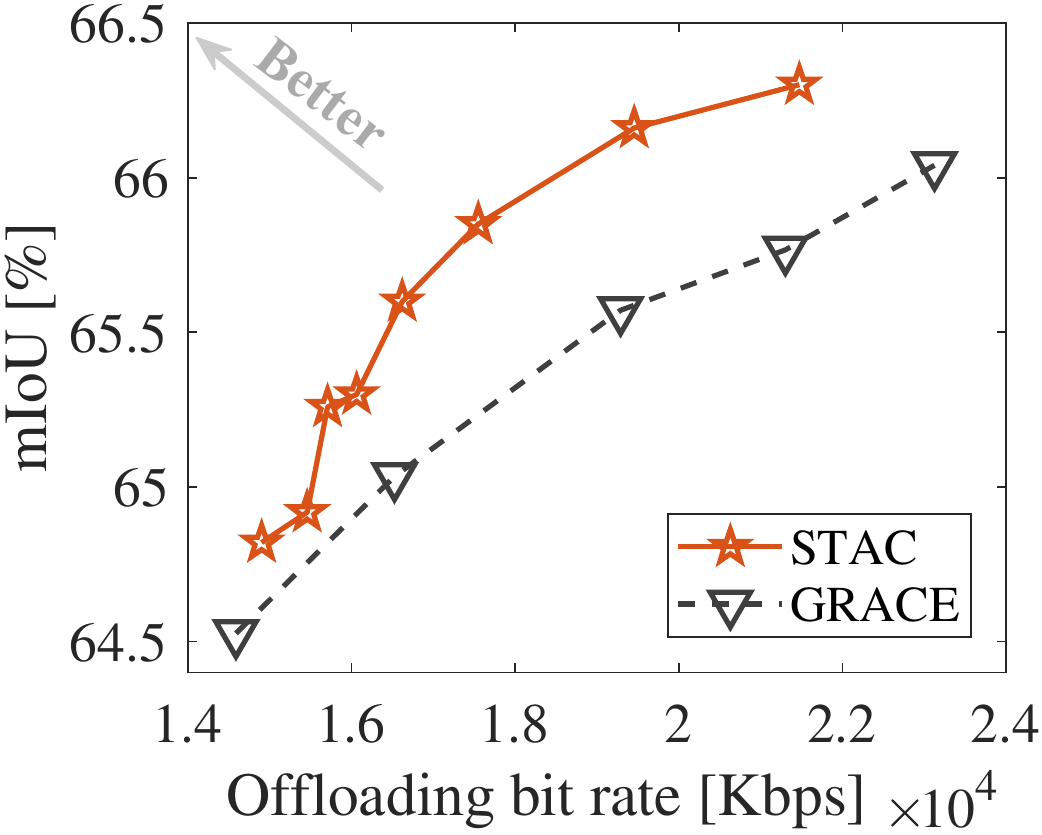"}\label{grace-D22}}
	\hfill
	\subfigure[vs. GRACE on DRN-D-38]{
		\includegraphics[width=0.48\linewidth]{"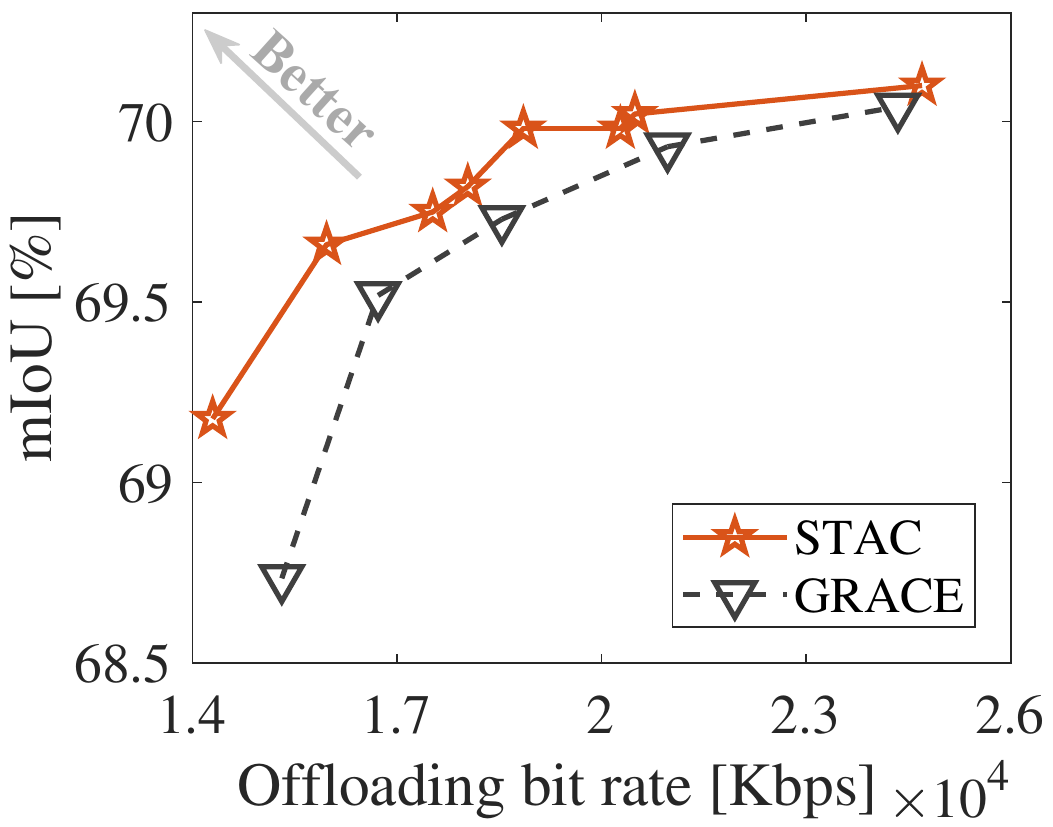"}\label{grace-D38}}\vspace{-0.15cm}
	\\
	\subfigure[vs. H.264 on DRN-D-22]{
		\includegraphics[width=0.48\linewidth]{"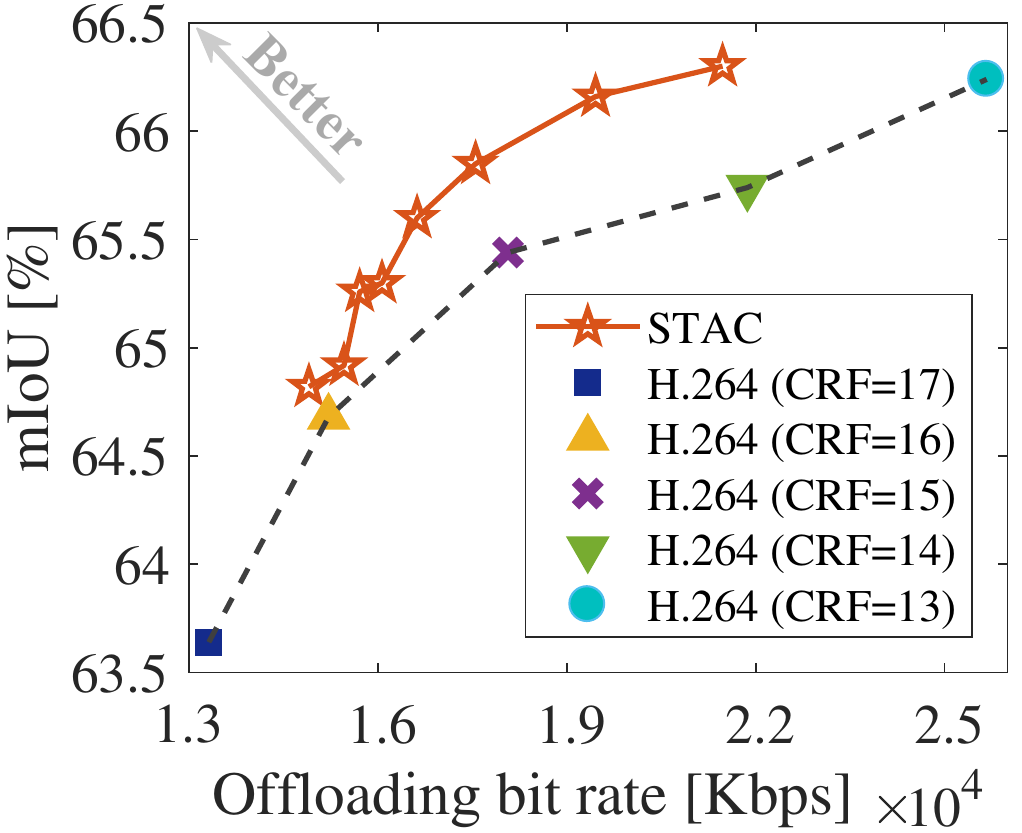"}\label{h264-D22}}
	\hfill
	\subfigure[vs. H.264 on DRN-D-38]{
		\includegraphics[width=0.48\linewidth]{"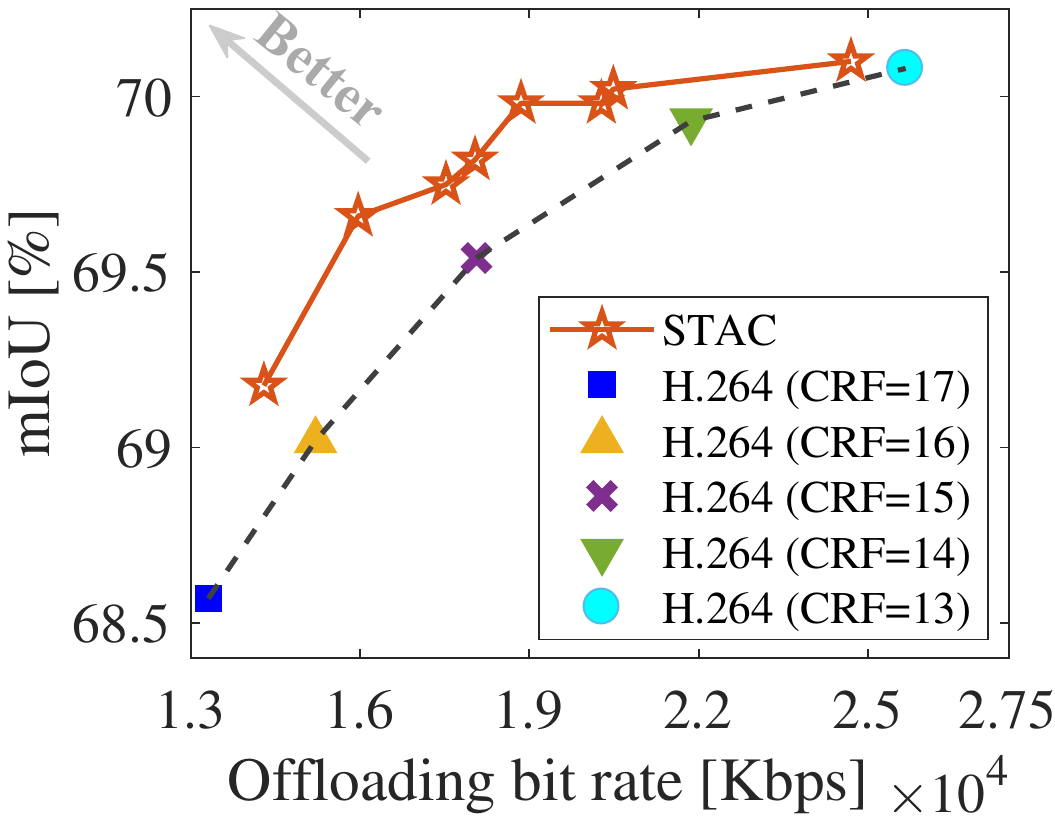"}\label{h264-D38}}
	\\
	\subfigure[vs. JPEG on DRN-D-22]{
		\includegraphics[width=0.48\linewidth]{"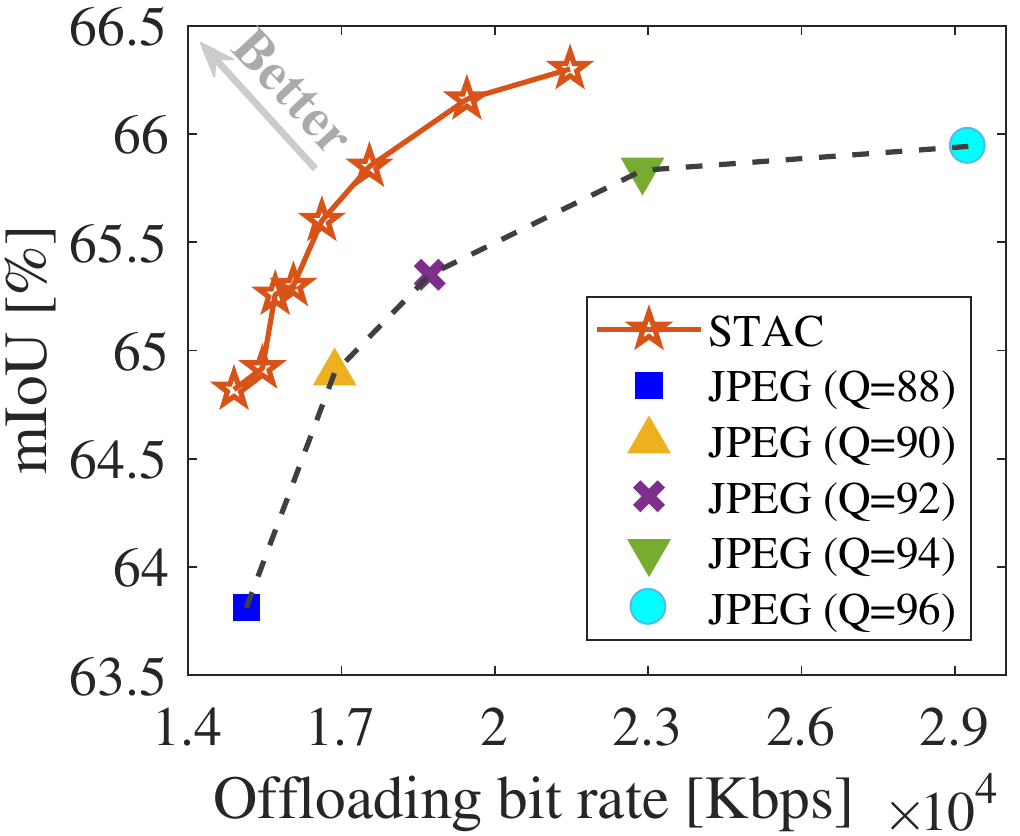"}\label{jpeg-D22}}
	\hfill
	\subfigure[vs. JPEG on DRN-D-38]{
		\includegraphics[width=0.48\linewidth]{"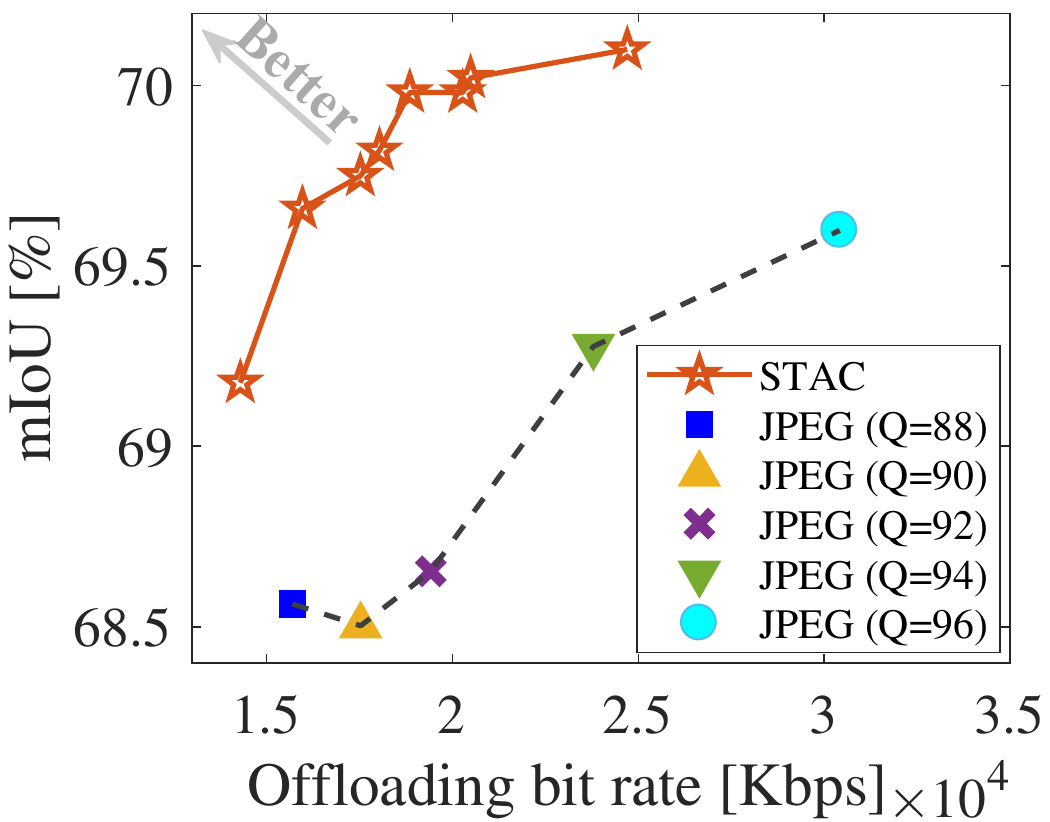"}\label{jpeg-D38}}\vspace{-0.1cm}
	\\
	
	\caption{Compare STAC with different compression algorithms/codecs.}\vspace{-0.3cm}
	\vspace{-0.1cm}\label{section2-vel-dis} 
\end{figure}

\begin{figure*}[t]
	\begin{minipage}[t]{0.325\linewidth}
		\centering
		\includegraphics[width=2.14in]{"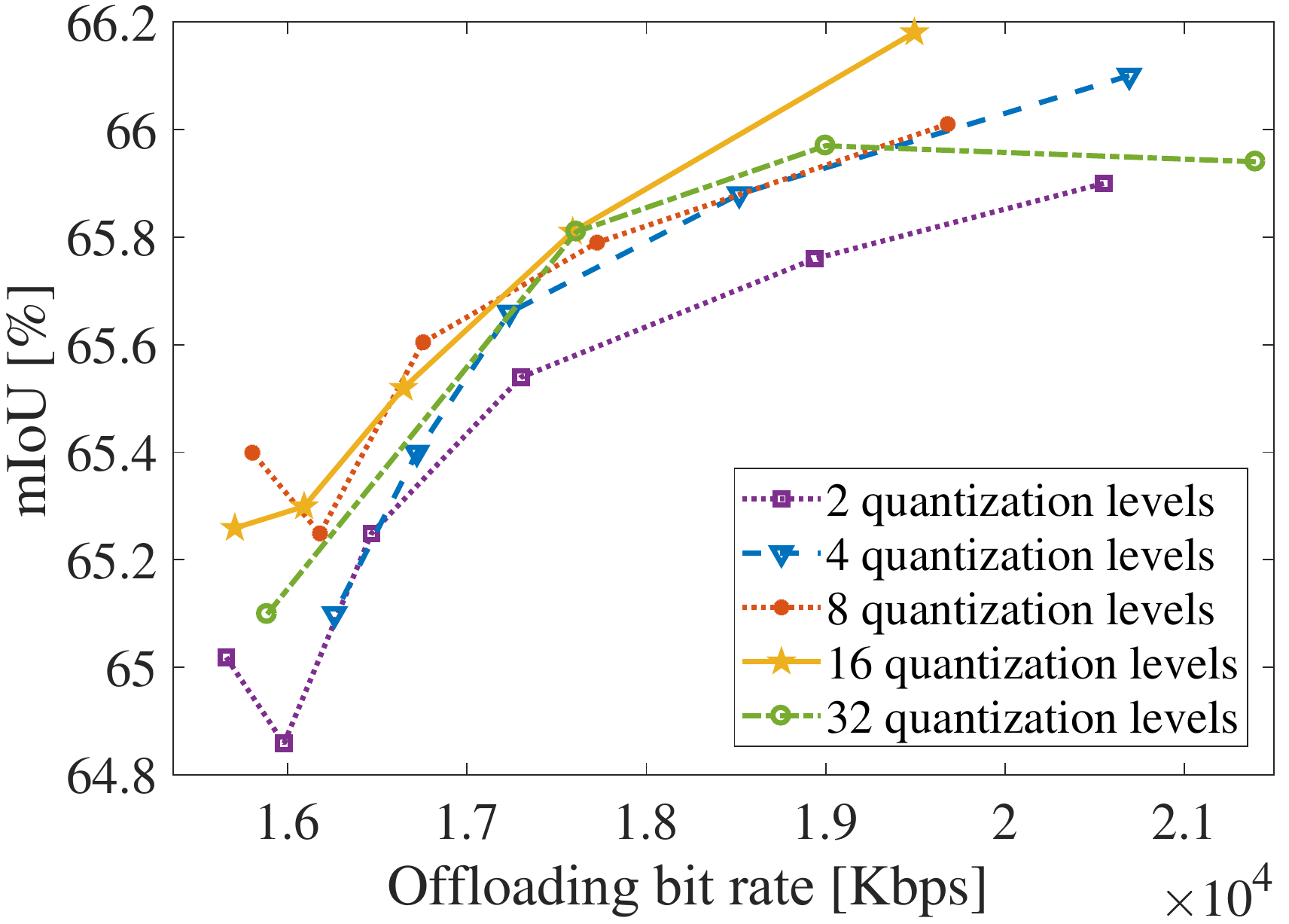"}
		\caption{The impact of different numbers of available quantization tables.}
		\label{levels}
	\end{minipage}%
	\begin{minipage}[t]{0.325\linewidth}
		\centering
		\includegraphics[width=2.1in]{"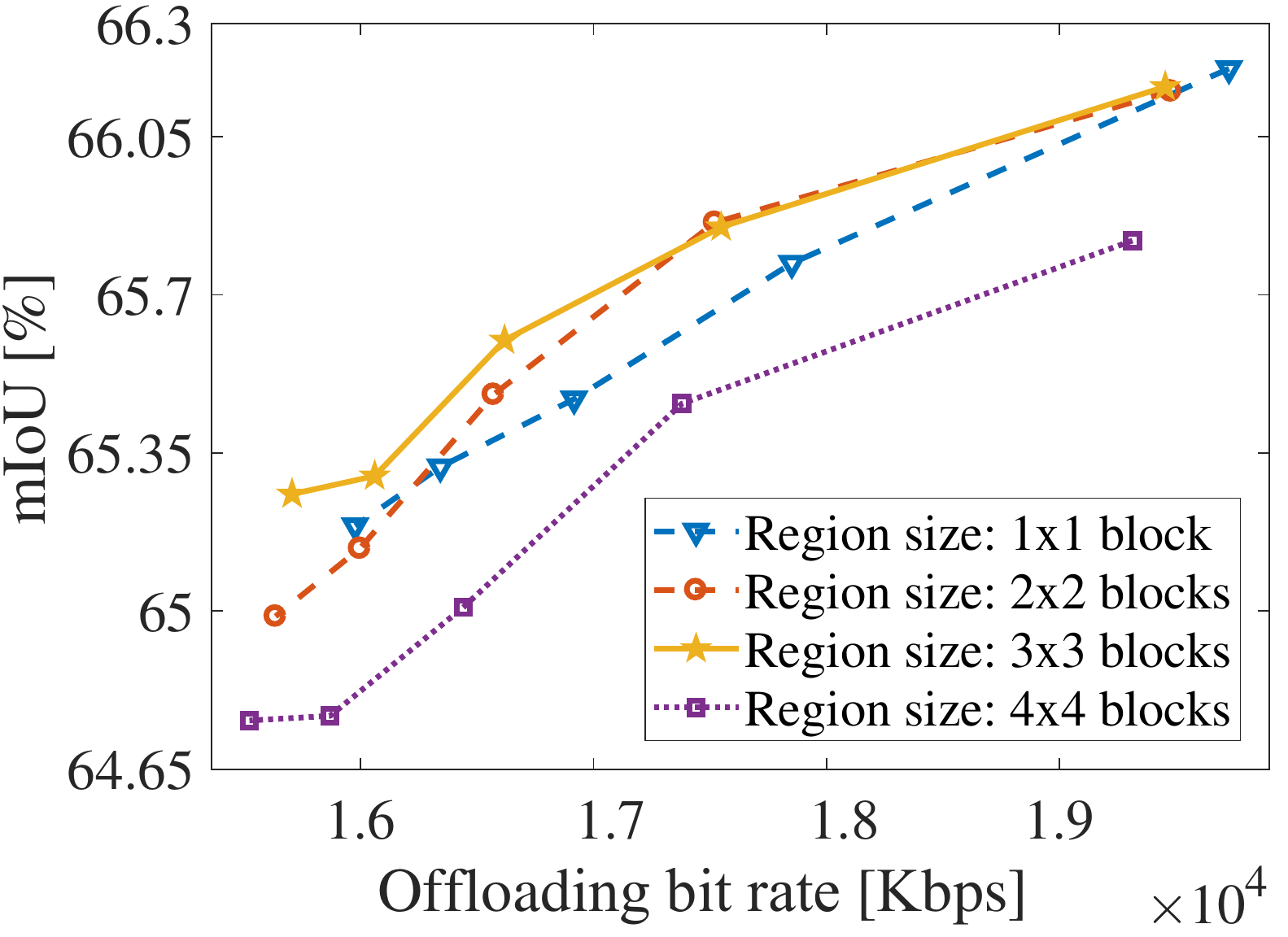"}
		\caption{The impact of region sizes.}\label{regions}
	\end{minipage}%
	\begin{minipage}[t]{0.325\linewidth}
		\centering
		\includegraphics[width=2.1in]{"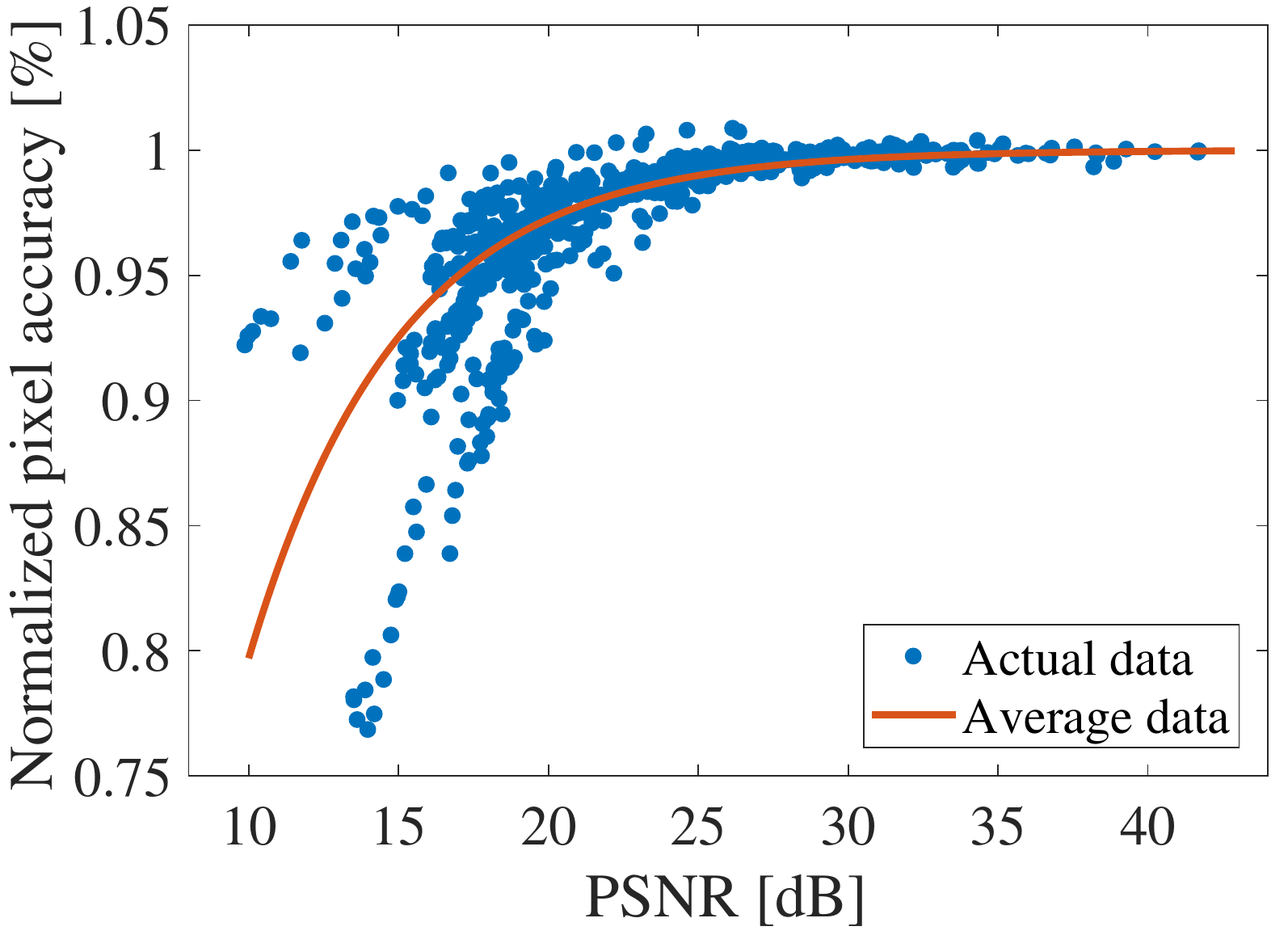"}
		\caption{Relationship between PSNR and semantic segmentation accuracy.}
		\label{psnr_acc}
	\end{minipage}\vspace{-0.4cm}
\end{figure*}


	\subsection{Comparison with Different Algorithms}\label{compare}
	
	Following the experimental setup above, we compare STAC with the baseline algorithms. 
	Table~\ref{table1} demonstrates the side-by-side comparison results on DRN-D-22 model. 
	To make the comparison more intuitive, we adjust the configuration of each algorithm to be at the same level of accuracy, and then the offloading bit rate indicates the  performance. We can notice that STAC reduces offloading bit rate (Kbps) by 15.87\% compared to GRACE, and the reduction reaches 24.23\% and 33.53\% compared to H.264 (CRF=13) and JPEG (Q=96). 
	Due to the temporal adaptive scheme, STAC can offload much fewer frames per second than H.264, which greatly reduces the bandwidth consumption. Furthermore, since the offloaded frames per second are the same for STAC, GRACE and JPEG, the reduction in offloading bit rate illustrates the effectiveness of spatial adaptive compression strategy.
	
	We further compare STAC with these algorithms under different configurations.

	\textbf{STAC vs. GRACE.} Both GRACE and STAC leverage the upperbound $B$ to adjust compression ratio and accuracy. We also set the same PSNR threshold of 26~dB to trigger the offloading of keyframes. Fig.~\ref{grace-D22} and Fig.~\ref{grace-D38} show the comparison results. 
	At first, we can observe that STAC outperforms GRACE in both accuracy and bandwidth consumption.
	When the accuracy is the same, STAC can save up to 20.95\% and 12.73\% of bandwidth consumption on DRN-D-22 and DRN-D-38, respectively. Similarly, when the offloading bit rate is the same, STAC can increase the accuracy (mIoU) by 0.17-0.7\% and 0.09\%-0.77\% on DRN-D-22 and DRN-D-38, respectively. It is noteworthy that this performance improvement entirely comes from the spatial adaptive strategy, which proves its effectiveness at different compression levels.
	
	

	\textbf{STAC vs. H.264.} Since Cityscapes dataset only provides the images of videos, we first adopt FFMPEG to merge these images into H.264 videos and adjust video qualities by constant rate factor (CRF). 
	Then, we split these videos into compressed images. The experimental results are shown in Fig.~\ref{h264-D22} and Fig.~\ref{h264-D38}. When accuracy is the same, STAC achieves a significant reduction in bandwidth consumption, reaching 24.23\% (CRF = 13) on DRN-D-22 and 15.41\% (CRF = 14) on DRN-D-38. Compared to H.264, STAC only needs to offload 6.8 frames per second in average while improving accuracy by an average of 0.48\% and 0.28\% on two DNN models respectively. 
	These results demonstrate the superiority of STAC in a bandwidth-limited environment. 
	
	

	\textbf{STAC vs. JPEG.} We further compare STAC with JPEG codec under different quality levels (Q), as we prototype STAC on JPEG. The experimental results are presented in Fig.~\ref{jpeg-D22} and Fig.~\ref{jpeg-D38}. Compared to JPEG, STAC achieves a significant improvement in accuracy, reaching 1.02\% (Q = 88) and 1.33\% (Q=92) on the two DNN models respectively, or a noticeable reduction in offloading bit rate, reaching 33.53\% (Q = 96) and 47.48\% (Q = 96). The reason for the performance improvement is twofold: (1) STAC incorporates the advantage of GRACE by leveraging the DNN's sensitivity w.r.t. frequency to generate different quantization table levels; (2) STAC generates spatial adaptive compression strategy online to maximize the accuracy or minimize the offloading bit rate.

	\subsection{Impact of Different Configurations} \label{configurations}
	In this subsection, we conduct several experiments to test different parameters and find the optimal settings.

	\textbf{Adjusting the number of offline quantization table levels.} 
	Theoretically, the more quantization table levels $\{l\}_{l = 1}^L $ available, the higher the accuracy but the heavier the communication load of compression strategy feedback and offloading. The reason is that we need more bits to transmit the levels. Thus, we evaluate STAC under different $L$ to find an appropriate value that can achieve a balance. The results are shown in Fig.~\ref{levels}. 
	When the offloading bit rate is the same, the accuracy increases significantly with $L$ from 2 to 16, but begins to drop as $L$ increases to 32. 
	The reason is that when the coverage and granularity of quantization tables are enough,
	the gain from increasing quantization levels becomes very small. Instead, the growing communication load leads to poor performance.  
	Thus, we set $L$ to 16, a value that allows STAC to achieve better accuracy while leading to a small offloading bit rate.

	\textbf{Adjusting the size of regions.}
	Theoretically, the smaller the size of regions, the higher the accuracy but the larger the data size required for compression strategy feedback and offloading. 
	Thus, we evaluate STAC under different region sizes, i.e., containing different numbers of blocks, such as $1\times1$ block, $2\times2$ blocks, etc. Since we test the impact of region sizes, the block size is configured to $8\times8$ to avoid repetition of experiments. The results are shown in Fig.~\ref{regions}. 
	From $1\times 1$ to $3\times 3$ blocks, the performance of STAC gradually increases, but drops significantly when the region size reaches $4\times 4$ blocks. 
	This is because, when the region size is large, obvious spatial sensitivity variation appears in the region. In contrast, the spatial sensitivity variation is negligible when the region size is small. Therefore, we set the region size to $3\times 3$ blocks to reduce the offloading bit rate without compromising accuracy.
	
	\textbf{Adjusting threshold for adaptive offloading.}
    STAC computes the similarity, i.e., PSNR between the propagated frame and the actual frame to infer the impact of propagation on segmentation accuracy. However, the relationship between PSNR and semantic segmentation accuracy is under-explored. 
    We, therefore, test semantic segmentation accuracy vs. PSNR when propagating across different numbers of frames. The results in Fig.~\ref{psnr_acc} shows a clear turning point around PSNR of 25~dB. When PSNR is higher than this value, the drop in semantic segmentation accuracy is small. However, when PSNR decreases, the semantic segmentation accuracy falls rapidly. Therefore, we set a PSNR  threshold as 26~dB to trigger the adaptive offloading mechanism, which minimizes offloading bit rate while maintaining a high level of accuracy.
    
\begin{figure}[t]
	\centerline{\includegraphics[width=0.45\textwidth]{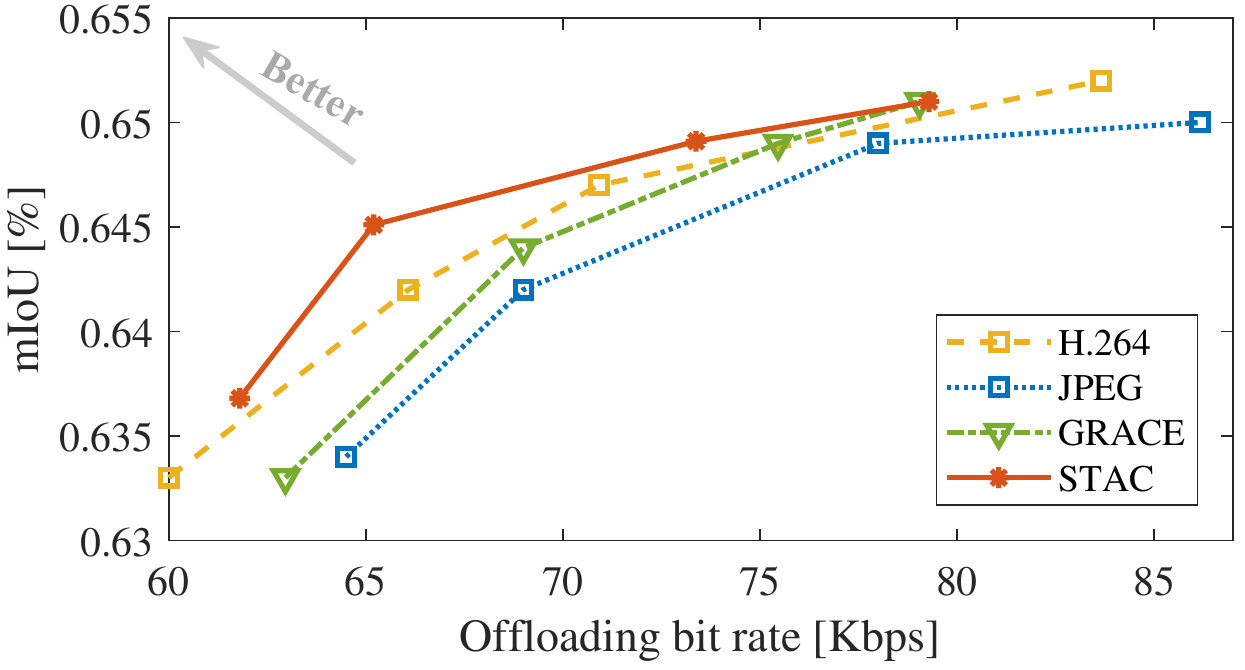}}
	\vspace{-0.2cm}\caption{The performance of different algorithms/codecs on BiSeNet model and CamVid dataset. }\vspace{-0.1cm}
	\label{camvid}\vspace{-0.3cm}
\end{figure}

	\subsection{Performance on Different Dataset}
	All previous experiments are based on the Cityscapes dataset. We further conduct experiments on a different dataset CamVid~\cite{brostow2009semantic} to test the adaptability of STAC. We use the BiSeNet~\cite{yu2018bisenet} model to perform semantic segmentation.
	The results are shown in Fig.~\ref{camvid}. It is shown that STAC outperforms other baseline algorithms in both accuracy and bandwidth consumption. 
	Compared to GRACE, STAC can still reduce up to 11.63\% of bandwidth consumption with comparable accuracy.
	However, as the frame rate of the videos in BiSeNet dataset is only 1 fps, this low temporal correlation cannot ensure the effectiveness of the temporal adaptive scheme in STAC. In other words, more keyframes are offloaded to accommodate the rapidly changing video content. As a result, the performance gap between STAC and H.264 is reduced.
	

	

	\section{Related Work} \label{related}
	\subsection{Semantic Segmentation}
	Existing DNN-based semantic segmentation can be summarized into two streams: semantic image segmentation~\cite{chen2018encoder,kirillov2019panoptic} and semantic video segmentation~\cite{paul2020efficient,zhu2019improving,xie2019source,shelhamer2016clockwork,mahasseni2017budget,li2018low,xu2018dynamic}. 
	For semantic video segmentation, the most straightforward way is to simply run semantic image segmentation on each frame~\cite{xie2019source} or keyframes~\cite{xu2018dynamic,paul2020efficient,zhu2019improving}. DNN models that take consecutive frames as input~\cite{mahasseni2017budget,shelhamer2016clockwork,li2018low} are trying to leverage temporal correlation to alleviate the excessive computational overhead.
	For example, Mahasseni et al.~\cite{mahasseni2017budget} leverage LSTM network and  Clockwork~\cite{shelhamer2016clockwork} share DNN features between frames.
	Recent works~\cite{xu2018dynamic,paul2020efficient,zhu2019improving} extract dense optical flow to propagate image segmentation results from keyframes to the target frame. 
	Our design also follows a similar framework, i.e., applying the compression strategy to offloaded keyframes for DNN inference and propagating segmentation results to other frames.
	
	\subsection{Compression Algorithms for Edge}
	Recent years have witnessed the prevalence of task-oriented compression algorithms. Existing compression algorithms~\cite{liu2019edge,du2020server,pakha2018reinventing,wang2021edgeduet} for object detection tasks typically generate bounding boxes of target objects and offload images within bounding boxes with high resolutions. 
	However, these compression algorithms are a poor fit for semantic segmentation, which does not have clear and highly concentrated RoIs. The state-of-the-art compression algorithm for semantic segmentation, GRACE~\cite{xie2019source} exploits the DNN's gradients (i.e., sensitivity) to different frequency- and color- components to optimize the quantization table. However, the same quantization table is applied uniformly to all blocks in an image, ignoring the nonuniform spatial sensitivity. Runespoor~\cite{wang2021enabling} leverages the inference accuracy inequality of different semantic classes to design compression and super resolution scheme. 
	Several studies~\cite{nakanoya2020task,hu2020starfish} have constructed autoencoders for detection tasks. However, the encoder is still equivalent to a fixed compression strategy without adaptation in spatial and temporal domains. 
	To reduce the bandwidth consumption, several works~\cite{chen2015glimpse,ran2018deepdecision} design frameworks to decide whether to offload inference tasks or do them locally. Similarly, we also design an adaptive offloading mechanism for STAC, which guarantees the accuracy while minimizing offloading bit rate.
	
	
	
	%
	\section{Conclusion} \label{conclusion}
	This paper presents STAC, which takes the first step to exploit DNN's gradients as spatial sensitivity metrics to achieve spatiotemporal adaptive compression. 
	STAC is tailored for edge-assisted semantic video segmentation and shows challenges posed in enabling compression to be adaptive to videos. We tackle these challenges through inter-frame propagation of compression strategies through dense optical flow.
	Experimental results demonstrate STAC achieving up to 20.95\% reduction in bandwidth consumption or 0.7\% increase in accuracy (mIoU), compared to the state-of-the-art algorithm. We believe STAC represents a significant step towards accelerating the broad application of semantics-aware mobile vision.
	

	

\section*{Acknowledgement}
This work was supported in part by the National Key R\&D Program of China under Grant 2020YFB1806600, National Science Foundation of China with Grant 62071194, Tencent Rhino-Bird Focus Research Project of Basic Platform Technology 2021, Science and Technology Department of Hubei Province with Grant 2021EHB002, RGC under Contract CERG 16204418, Contract 16203719, Contract 16204820, and Contract R8015; and in part by the Guangdong Natural Science Foundation under Grant 2017A030312008, the European Union's Horizon 2020 research and innovation programme under the Marie Skłodowska-Curie grant agreement No 101022280.
 
\balance

\bibliographystyle{unsrt}
\bibliography{IEEEabrv,./STAC-SS}

\begin{thebibliography}{10}

\bibitem{cars}
Manouchehr Rafie.
\newblock Autonomous vehicles: {AI} must accelerate.
\newblock Accessed July 12, 2021.

\bibitem{google}
Christine Chan.
\newblock How to use portrait mode in facetime on iphone and ipad.
\newblock Accessed Jun 24, 2021.

\bibitem{youtube}
Anil Chandra~Naidu Matcha.
\newblock A 2021 guide to semantic segmentation.
\newblock Accessed May 24, 2021.

\bibitem{try-on}
Liz Flora.
\newblock Google and {S}napchat turn to beauty tech for {AR} makeup try-on.
\newblock Accessed JAN 06, 2021.

\bibitem{xie2019source}
Xiufeng Xie and Kyu-Han Kim.
\newblock Source compression with bounded {DNN} perception loss for {IoT} edge
  computer vision.
\newblock In {\em The 25th Annual International Conference on Mobile Computing
  and Networking}, pages 1--16, 2019.

\bibitem{wallace1992jpeg}
Gregory~K Wallace.
\newblock The {JPEG} still picture compression standard.
\newblock {\em IEEE transactions on consumer electronics}, 38(1):xviii--xxxiv,
  1992.

\bibitem{wiegand2003overview}
Thomas Wiegand, Gary~J Sullivan, Gisle Bjontegaard, and Ajay Luthra.
\newblock Overview of the {H}. 264/{AVC} video coding standard.
\newblock {\em IEEE Transactions on circuits and systems for video technology},
  13(7):560--576, 2003.

\bibitem{kroeger2016fast}
Till Kroeger, Radu Timofte, Dengxin Dai, and Luc Van~Gool.
\newblock Fast optical flow using dense inverse search.
\newblock In {\em European Conference on Computer Vision}, pages 471--488.
  Springer, 2016.

\bibitem{mahasseni2017budget}
Behrooz Mahasseni, Sinisa Todorovic, and Alan Fern.
\newblock Budget-aware deep semantic video segmentation.
\newblock In {\em Proceedings of the IEEE Conference on Computer Vision and
  Pattern Recognition}, pages 1029--1038, 2017.

\bibitem{shelhamer2016clockwork}
Evan Shelhamer, Kate Rakelly, Judy Hoffman, and Trevor Darrell.
\newblock Clockwork convnets for video semantic segmentation.
\newblock In {\em European Conference on Computer Vision}, pages 852--868.
  Springer, 2016.

\bibitem{li2018low}
Yule Li, Jianping Shi, and Dahua Lin.
\newblock Low-latency video semantic segmentation.
\newblock In {\em Proceedings of the IEEE Conference on Computer Vision and
  Pattern Recognition}, pages 5997--6005, 2018.

\bibitem{xu2018dynamic}
Yu-Syuan Xu, Tsu-Jui Fu, Hsuan-Kung Yang, and Chun-Yi Lee.
\newblock Dynamic video segmentation network.
\newblock In {\em Proceedings of the IEEE conference on computer vision and
  pattern recognition}, pages 6556--6565, 2018.

\bibitem{paul2020efficient}
Matthieu Paul, Christoph Mayer, Luc~Van Gool, and Radu Timofte.
\newblock Efficient video semantic segmentation with labels propagation and
  refinement.
\newblock In {\em Proceedings of the IEEE/CVF Winter Conference on Applications
  of Computer Vision}, pages 2873--2882, 2020.

\bibitem{zhu2019improving}
Yi~Zhu, Karan Sapra, Fitsum~A Reda, Kevin~J Shih, Shawn Newsam, Andrew Tao, and
  Bryan Catanzaro.
\newblock Improving semantic segmentation via video propagation and label
  relaxation.
\newblock In {\em Proceedings of the IEEE/CVF Conference on Computer Vision and
  Pattern Recognition}, pages 8856--8865, 2019.

\bibitem{ehrlich2020quantization}
Max Ehrlich, Larry Davis, Ser-Nam Lim, and Abhinav Shrivastava.
\newblock Quantization guided jpeg artifact correction.
\newblock In {\em Computer Vision--ECCV 2020: 16th European Conference,
  Glasgow, UK, August 23--28, 2020, Proceedings, Part VIII 16}, pages 293--309.
  Springer, 2020.

\bibitem{cordts2016cityscapes}
Marius Cordts, Mohamed Omran, Sebastian Ramos, Timo Rehfeld, Markus Enzweiler,
  Rodrigo Benenson, Uwe Franke, Stefan Roth, and Bernt Schiele.
\newblock The cityscapes dataset for semantic urban scene understanding.
\newblock In {\em Proceedings of the IEEE conference on computer vision and
  pattern recognition}, pages 3213--3223, 2016.

\bibitem{brostow2009semantic}
Gabriel~J Brostow, Julien Fauqueur, and Roberto Cipolla.
\newblock Semantic object classes in video: A high-definition ground truth
  database.
\newblock {\em Pattern Recognition Letters}, 30(2):88--97, 2009.

\bibitem{DRN22}
Fisher Yu.
\newblock Dilated residual networks. https://github.com/fyu/drn, 2018.

\bibitem{yu2017dilated}
Fisher Yu, Vladlen Koltun, and Thomas Funkhouser.
\newblock Dilated residual networks.
\newblock In {\em Proceedings of the IEEE conference on computer vision and
  pattern recognition}, pages 472--480, 2017.

\bibitem{yu2018bisenet}
Changqian Yu, Jingbo Wang, Chao Peng, Changxin Gao, Gang Yu, and Nong Sang.
\newblock Bisenet: Bilateral segmentation network for real-time semantic
  segmentation.
\newblock In {\em Proceedings of the European conference on computer vision
  (ECCV)}, pages 325--341, 2018.

\bibitem{chen2018encoder}
Liang-Chieh Chen, Yukun Zhu, George Papandreou, Florian Schroff, and Hartwig
  Adam.
\newblock Encoder-decoder with atrous separable convolution for semantic image
  segmentation.
\newblock In {\em Proceedings of the European conference on computer vision
  (ECCV)}, pages 801--818, 2018.

\bibitem{kirillov2019panoptic}
Alexander Kirillov, Ross Girshick, Kaiming He, and Piotr Doll{\'a}r.
\newblock Panoptic feature pyramid networks.
\newblock In {\em Proceedings of the IEEE/CVF Conference on Computer Vision and
  Pattern Recognition}, pages 6399--6408, 2019.

\bibitem{liu2019edge}
Luyang Liu, Hongyu Li, and Marco Gruteser.
\newblock Edge assisted real-time object detection for mobile augmented
  reality.
\newblock In {\em The 25th Annual International Conference on Mobile Computing
  and Networking}, pages 1--16, 2019.

\bibitem{du2020server}
Kuntai Du, Ahsan Pervaiz, Xin Yuan, Aakanksha Chowdhery, Qizheng Zhang, Henry
  Hoffmann, and Junchen Jiang.
\newblock Server-driven video streaming for deep learning inference.
\newblock In {\em Proceedings of the Annual conference of the ACM Special
  Interest Group on Data Communication on the applications, technologies,
  architectures, and protocols for computer communication}, pages 557--570,
  2020.

\bibitem{pakha2018reinventing}
Chrisma Pakha, Aakanksha Chowdhery, and Junchen Jiang.
\newblock Reinventing video streaming for distributed vision analytics.
\newblock In {\em 10th $\{$USENIX$\}$ Workshop on Hot Topics in Cloud Computing
  (HotCloud 18)}, 2018.

\bibitem{wang2021edgeduet}
Xu~Wang, Zheng Yang, Jiahang Wu, Yi~Zhao, and Zimu Zhou.
\newblock {EdgeDuet}: Tiling small object detection for edge assisted
  autonomous mobile vision.
\newblock In {\em IEEE INFOCOM 2021-IEEE Conference on Computer
  Communications}, pages 1--10. IEEE, 2021.

\bibitem{wang2021enabling}
Yiding Wang, Weiyan Wang, Duowen Liu, Xin Jin, Junchen Jiang, and Kai Chen.
\newblock Enabling edge-cloud video analytics for robotics applications.
\newblock In {\em Proceedings of the IEEE International Conference on Computer
  Communications, Virtual Conference}, pages 10--13, 2021.

\bibitem{nakanoya2020task}
Manabu Nakanoya, Sandeep Chinchali, Alexandros Anemogiannis, Akul Datta, Sachin
  Katti, and Marco Pavone.
\newblock Task-relevant representation learning for networked robotic
  perception.
\newblock {\em arXiv preprint arXiv:2011.03216}, 2020.

\bibitem{hu2020starfish}
Pan Hu, Junha Im, Zain Asgar, and Sachin Katti.
\newblock Starfish: resilient image compression for aiot cameras.
\newblock In {\em Proceedings of the 18th Conference on Embedded Networked
  Sensor Systems}, pages 395--408, 2020.

\bibitem{chen2015glimpse}
Tiffany Yu-Han Chen, Lenin Ravindranath, Shuo Deng, Paramvir Bahl, and Hari
  Balakrishnan.
\newblock Glimpse: Continuous, real-time object recognition on mobile devices.
\newblock In {\em Proceedings of the 13th ACM Conference on Embedded Networked
  Sensor Systems}, pages 155--168, 2015.

\bibitem{ran2018deepdecision}
Xukan Ran, Haolianz Chen, Xiaodan Zhu, Zhenming Liu, and Jiasi Chen.
\newblock Deepdecision: A mobile deep learning framework for edge video
  analytics.
\newblock In {\em IEEE INFOCOM 2018-IEEE Conference on Computer
  Communications}, pages 1421--1429. IEEE, 2018.

\end{thebibliography}

\end{document}